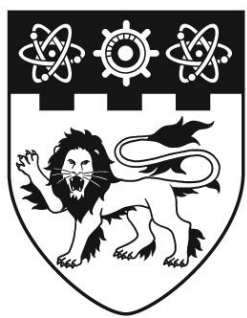

# Development of an Intuitive Foot-Machine Interface for Robotic Surgery

Huang Yanpei

School of Mechanical and Aerospace Engineering

2018

**Development of an Intuitive Foot-Machine Interface for Robotic Surgery**

Huang Yanpei

School of Mechanical and Aerospace Engineering

Confirmation of Candidature Report submitted to

the Nanyang Technological University

in partial fulfilment of the requirements for the degree of

Doctor of Philosophy

2018

# Contents

# List of Acronyms

| | |
|---|---|
| **AESOP** | **A**utomated **E**ndoscopic **S**ystem for **O**ptimal **P**ositioning |
| **DOF** | **D**egree **o**f **f**reedom |
| **FI** | **F**oot **I**nterface |
| **HMI** | **H**uman-**M**achine **I**nteraction |
| **LapMan** | **Lap**aroscope **M**anipulator |
| **LER** | **L**ight **E**ndoscope **R**obot |
| **LODEM** | **M**obile **L**ocally **O**perated **D**etachable **E**nd-effector **M**anipulator |
| **MB** | **M**oving **F**rame of the foot interface |
| **BB** | **B**ase **F**rame of the foot interface |
| **RPR** | **R**otation joint-**P**rismatic joint-**R**otation joint |
| **SGD** | **S**in**g**apore **D**ollars |
| **F** | **F**orward |
| **B** | **B**ackward |
| **TU** | **T**oe **U**p |
| **TD** | **T**oe **D**own |
| **S** | **S**upination (left torsion) |
| **P** | **P**ronation (right torsion) |
| **LF** | **L**eft & **F**orward |
| **RF** | **R**ight & **F**orward |
| **LB** | **L**eft & **B**ackward |
| **RB** | **R**ight & **B**ackward |
| **LTU** | **L**eft & **T**oe **U**p |
| **RTU** | **R**ight & **T**oe **U**p |
| **LTD** | **L**eft & **T**oe **D**own |
| **RTD** | **R**ight & **T**oe **D**own |
| **FTU** | **F**orward & **T**oe **U**p |
| **BTU** | **B**ackward **&** **T**oe **U**p |
| **FTD** | **F**orward **&** **T**oe **D**own |
| **BTD** | **B**ackward & **T**oe **D**own |
| **kNN** | **k**-**n**earest **n**eighbor's algorithm |
| **ICA** | **I**ndependent **C**omponent **A**nalysis |
| **OR** | **O**peration **R**oom |

# Lists of symbols

*x, y,* ∅   Cartesian parameters of the mobile plate of center point C

{O}, {C} BF and MF Cartesian coordinate frames.

a, b   BF length, width

a', b'   MF length, width

c   Distance between adjacent springs on the long side of BF

c'   Distance between adjacent springs on the moving plate

$l_{0i}$   $i$th spring guide initial length without input

$l_i$   $i$th spring guide lengthes with some random input

$A_i, B_i$   $i$th spring guide connection points with BF and MF point.

$F_{0i}$   $i$th spring initial compression force without external force

$F_i$   $i$th spring compression force

$k_i$   $i$th spring stiffness

$F_x$   Resultant force in y direction of MF

$F_y$   Resultant force in x direction of MF

$F$   Resultant force of MF on center point C

$M$   Torque on the MF around center point C

$F_F$   External force from human foot

$M_F$   External torque from human foot

$\alpha_i$   $i$th spring force direction angle with x axis, from x+, anti-clockwise is +, clockwise is –

$\beta_i$   angle between vector $\overrightarrow{CB_i}$ and $\overrightarrow{A_iB_i}$

# Lists of figures

## List of tables

# Abstract

The human-machine interface is of critical importance for master-slave control of robotic system for surgery, in which current systems offer the control or two robotic arms tele-operated by the surgeon's hands. To relax the need for surgical assistants and augment dexterity in surgery, it has been recently proposed to use a robot like a third arm that can be controlled seamlessly, independently from the natural arms, and work together with them. This report will develop and investigate this concept by implementing foot control of a robotic surgical arm. A novel passive haptic foot-machine interface system and analysis of its performances was introduced in this report. This interface using a parallel-serial hybrid structure with springs and force sensors, which allows intuitive control of a slave robotic arm with four degrees of freedom (dof). The elastic-isometric design enables a user to control the interface system accurately and adaptively, with an enlarged sensing range breaking physical restriction of the pedal size. A subject specific (independent component analysis, ICA) model is identified to map the surgeon's foot movements into kinematic parameters of the slave robotic arm. To validate the system and assess the performance it allows, 10 subjects carried out experiments to manipulate the foot-machine interface system in various movements. With these experimental data, the mapping models were built and verified. A comparison between different mapping models was made and analyzed proving the ICA algorithm is obviously dominant over other methods. In a last chapter I outline the further steps for my PhD project.

# Chapter 1 Introduction

## 1.1 Background and motivation

Robotic surgery is increasingly used in hospitals and will have important societal and economic impacts. It has been developed to improve keyhole surgery, in which the surgeon sits behind a console with high resolution 3D view, and tele-manipulates two robot arms inserted into patient's body using console as Figure 1 shows. During a surgical intervention, both of surgeon's hands are occupied with tools or hand consoles. Therefore, many operation procedures require the help of a surgeon assistant, for example to hold a tissue on which the surgeon will operate. However, the efficiency of surgical interventions can suffer from suboptimal cooperation between the surgeon and assistants [1]. In particular, communication errors are inevitable with novice assistants or in a new team [2]. Even in a good cooperation team, communication problems may still happen because of technical issue like different views (2D vs. 3D) in robotic surgery [3]. Any mistake in surgery may affect the patient's health.

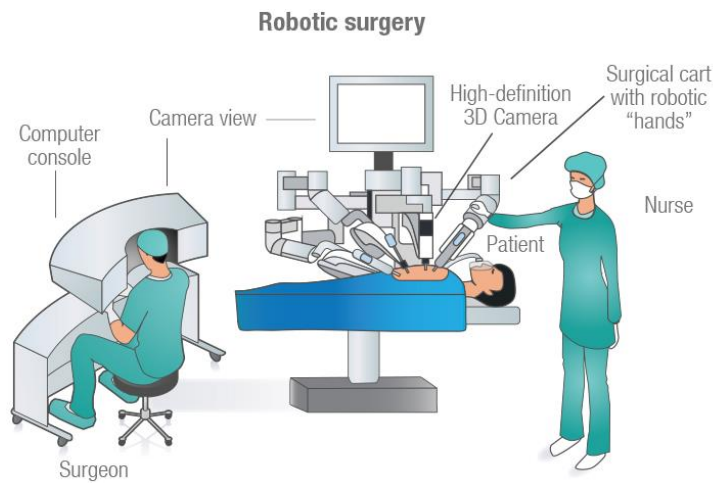

**Figure 1: Robotic surgery [4].**

One typical example is the laparoscopic surgery in which usually two or three surgical assistants [5] are needed. In the procedure, the primary surgeon cannot control the operative view himself/herself. There is an assistant to specially hold the laparoscopic camera. This indirect observing breaks the natural coordination of surgeon's hand and eye. The camera assistant need to coordinate well with surgeon and may suffer from fatigue in long lasting and repetitive actions. The tremor of assistant's hands lead to unstable image and hinder the operation of surgeon [6]. Several mechanical camera holders have been invented to facilitate the surgery. However, as the two hands of the surgeon were already occupied by main operation, the guidance on camera holder device is still conducted by an assistant. The communication problem may still occur in these cases [3].

Some passive camera positioners allow solo-operation by one surgeon, the surgical tasks are completed by hands sequentially which is time-consuming and cause distraction and/or interruption to current operation. For instance, when the surgeon wishes to change the current camera view in laparoscopic surgery, he or she needs to release the current surgical instruments, disengages the locking mechanism, operating the third instrument, reengages the locking mechanism, pick up previous instruments and resumes the procedures. There are also active camera holder which can free hands but using head movements [7] [8], voice control [9], finger [10] [11]and foot switches, as described below:

**Head control**: In the systems of EndoAssist [7] and Freehand [8], 4dof head movements (3 rotations and forward/backward translation) are detected using an infrared transmitter attached to the surgeon's head and a receiver installed on the monitor, which are used to control the endoscope holder robots. Figure 1 (a) shows how head movement controls a 4dof endoscope arm: The view orientation of the endoscope, i.e. up and down, left and right, and face rotation, follows the three head rotations, while the antero-posterior head movement controls the camera zoom in and out.

**Voice control**: The surgeon is typically using a headset with microphone to communicate commands using simple words like "up", "down", "right", "left", "in", "out", which are subsequently repeated it the headphones for checking that the proper command was identified by the system. The ZEUS (Figure 1(b)) and AESOP (Automated Endoscopic System for Optimal Positioning) 2000 [9] robots are using such voice control.

**Hand/Finger control**: Interfaces such as a joystick or switches are directly attached on the surgical instrument. For example, the Lapman [10] robot has a joystick clipped on the instrument under the surgeon's index finger that can be controlled using one finger without affecting the hand operation. The FIPS endoarm [11] equipped a finger-ring joystick which make the endoscope tip movement corresponding to fingertip movement.

**Eye gaze control**: Interfaces have been developed using eye gaze to detect the intention of the operator [5-7]. In this system, an eye tracker device is needed to infer where the surgeon is looking at by monitoring his/her eye movements. The laparoscope camera could be controlled by following the user's gaze position. For example, Noonan et al. have proposed a system to control a robotic laparoscope movements through surgeon's gaze on a button or particular screen portions [12]. Staub et al.'s system [13] also uses gaze to control laparoscope moving directions with a pedal to activate the robot. The Senhance surgical system [14] of TransEnterix Surgical Inc. could tracks surgeon's eye movements to control the camera fields.

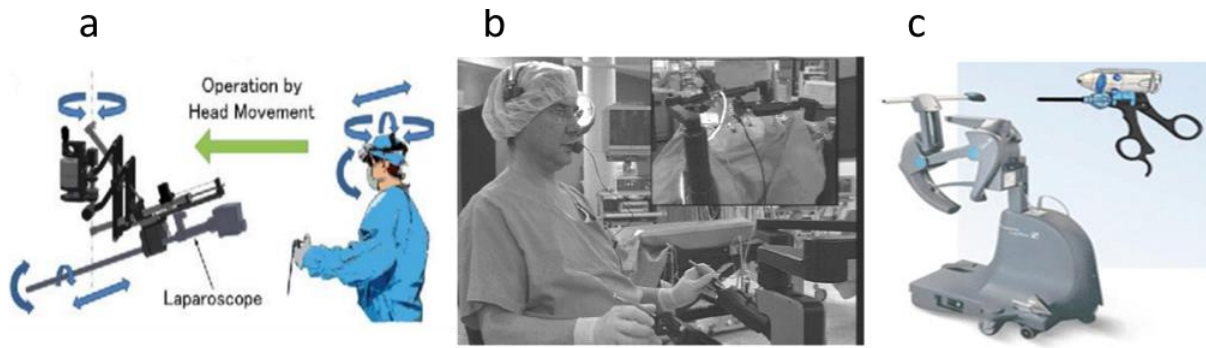

**Figure 2: (a) Head control (b) Voice control of ZEUS robot (c) Finger control of LapMan by Medsys, Gembloux, BELGIUM.**

" Hand-free" control such as with the tongue [15][16] is also used to replace missing functions such as to control an upper limb prosthesis. For instance, D. Johansen et al. [16]developed an inductive tongue control system for controlling prosthetic hands. A special activation unit was glued to the tip of tongue (Figure 2(a)), which can touch inductive sensors hosted in the upper palatal (Figure 2(b)) with five activated areas used to control a prosthetic hand.

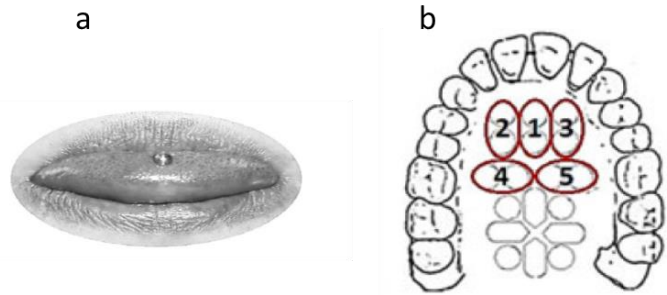

**Figure 3: Tongue control with (a) activation unit glued to the tip of the tongue and (b) layout of activation areas on the upper palatal.**

While these control strategies could be used to free the surgeon's hands throughout the operation, all suffer from limitations in specific applications. Voice recognition may fail due to environmental noise or other surgeons nearby, and yields so far only a limited vocabulary of commands. Controlling head movement may distract the surgeon's attention, who should be able to use natural eye gazing and head to coordinate whole body movements [17]. Inadvertent negligence may cause security issues, thereby some head control systems would still need additional activation buttons or pedals to distinguish control from casual movements. Finger control is not completely hands-free, the operation may be affected by the hands operation. Tongue control cannot be used simultaneously to speaking, which is necessary to interact with the team during surgical interventions. Those interfaces may less intuitive and slower, as a controlling command can only move one dof one time[5]. And the controlling operation may interrupt the main procedure of hands [18], e.g. rotating head from interested field.

Therefore, there is a need for a hand-free interface which allows the surgeon to continuously control the movements of a robotic arm in an intuitive and effective way. Recent studies [19][20] have demonstrated the efficient coordination and superior manipulation capabilities of foot cooperating with hands enabling the "three hands surgery". However, it is still an open question, how to use foot to accurately control a surgical robot. This project focuses on developing a foot interface and related methodology which can control multi-dof of a surgical tool in an intuitive and effective way using foot movements and haptic feedback.

## 1.2 Objective and scope

The goal of this project is to develop a foot-control system with build-in algorithm and methodology which can best reflect operators' intention to effectively and intuitively remote control a robotic surgical tool. This system can address typical limitations of current interfaces, and offer features that improve the performance of users. Then, the scope of this project includes:

(1) Designing a foot interface which could control at least four dof of a robot. The interface should have the ability to (*i*) capture and identify natural foot motions; (*ii*) provide sufficient haptic feedback;(*iii*) enable ergonomic and comfort operation.

(2) As foot is less accurate compared to hand operation. An effective algorithm/mapping model should be developed between user and the interface which can best reflect user's control intention compensating the inaccurate operation of the user's foot.

(3) The whole foot interface system is a teleoperation system which should include the slave robot with the foot interface acting as a master device. The communication and control mechanism should be well established to make the foot tele-control as transparent as possible.

(4) A serial of experiments will be designed to conduct the user study investigating on human learning and manipulating capability using foot with the developed foot interface and also the state of art foot interface.

(5) The system will finally be equipped with a complete methodology and optimized structure based on simulation and user study.

## 1.3 Report outline

This report has five chapters in total. Chapter1 has introduced some background of the robotic surgery and its communication error, some hand-free interfaces were reviewed and their weakness motivate the current study. Based on some studies on foot [19][20], one can believe that foot control is a source with great potential to intuitively and effectively control a robot in surgery.

Thus, the objective of the project is established with developing the most intuitive foot interface to control a surgical robot. The designed foot interface should adapt to the characteristics of the user on foot biomechanics, foot capability and ergonomic requirements which are described in Chapter 2. The available foot interfaces in robotic surgery are also reviewed and discussed in their strengths and weaknesses. Chapter 3 will explain the design process for the foot interface including mechanical structure and kinematics calculation. The prototype interface was built and used to study the human operation movements characteristics in the experiment describe in Chapter 4. Based on the study, an algorithm to extract operator specific commands from human natural movement patterns was established which enabling intuitive and efficient control of a slave mechanism. The developed interface will also be compared with the existing interface and tested in clinical applications, the intuitive and natural interaction and control strategy between human and surgical robot will be explored in the further work (Chapter 5).

# Chapter 2 Literature review

To design an intuitive human machine interface, the user's characteristics and the interaction between user and the interface system are key issues need to be analyzed first. The first part of this chapter will go through the biomechanics of the human foot and leg, identify the foot machine interaction capability and ergonomic requirements that foot interface design should consider. The later part will review the representative foot interfaces in robotic surgery.

## 2.1 Characteristics of users

### 2.1.1 Foot biomechanics

The foot is a complex structure which contains 26 bones, 2 sesamoid bones, 33 joints, 19 muscles and 107 ligaments[21]. It is a highly dexterous system with advanced movement using multiple joints. Each joint provides multiple movements with varying ranges.

#### 2.1.1.1 Ankle joint

Ankle joint is the joint connecting leg and foot. It is not one joint but three separate articulations which are subtalar joint, tibiofibular joint and talocrural joint enabling three rotations with each in two directions. Those motions are described below and the typical range of motion of ankle are shown in table 1.

Table 1: Typical range of motion of human ankle (Maximum allowable motion) [24].

| Motion | Range (deg) | Mean(deg) | Standard deviation(deg) |
| --- | --- | --- | --- |
| Dorsiflexion | 20.3–29.8 | 24.68 | 3.25 |
| Plantarflexion | 37.6–45.75 | 40.92 | 4.32 |
| Inversion | 14.5–22 | 16.29 | 3.88 |
| Eversion | 10–17 | 15.87 | 4.45 |
| Adduction | 22–36 | 29.83 | 7.56 |
| Abduction | 15.4–25.9 | 22.03 | 5.99 |

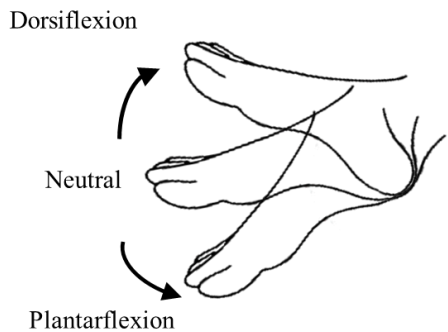

**Figure 4: Dorsiflexion and plantarflexion of ankle in frontal plane**

- Dorsiflexion and plantarflexion in frontal plane (Figure 4): Plantarflexion is the movement of foot away from the leg (moving upwards). Dorsiflexion is the movement of foot close to the leg (moving downward). These movements are commonly used foot control gesture used to operate pedal. The pedal can be toe activated or heel activated depending on where is the foot anchored [22].

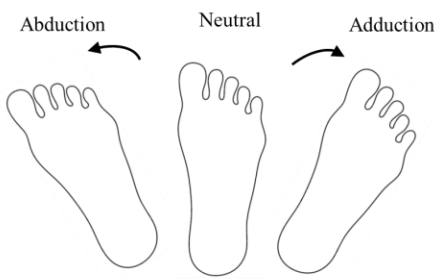

**Figure 5: Abduction and adduction of ankle in sagittal plane (Top view of foot).**

- Abduction and adduction in sagittal plane (Figure 5): Abduction is the movement of the foot away from the centerline of the body (moving outwards). Adduction is the horizontally movement of foot toward the centerline of the body (moving inwards). As a gesture, these movements are interpreted as heel rotation when pivoting around the heel or as toe rotation if pivoting round the toe [22]. In the interface of Foot Menu [23], the user was using abduction and adduction around heel to control the horizontal movement of the cursor.

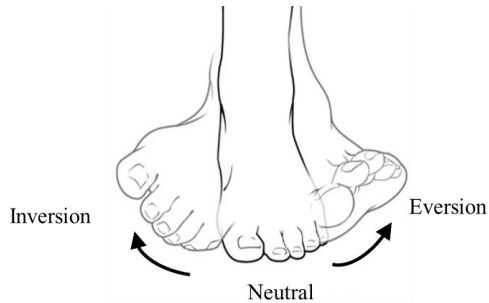

**Figure 6: Eversion and inversion of ankle in transverse plane.**

- Eversion and inversion in transverse plane: A foot twist outwards (Eversion) and twist inwards (Inversion). The range of motion for along this axis is very limited. Those movements are often combined with the other rotations into supination and pronation which are introduced below.

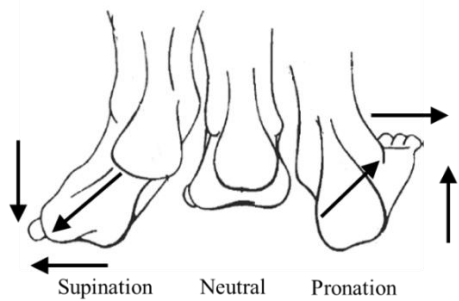

**Figure 7: Pronation and Supination of ankle.**

- Pronation and Supination: These are tri-planar movements that combines these movements. Pronation is the upward and outward movement of the foot involving dorsiflexion, abduction and eversion. Supination is a downward and inward movement of the foot involving plantarflexion, adduction and inversion. They are typically used to move foot joysticks.

**2.1.1.2 Knee joint**

The knee is a hinge joint enable relative movement between upper part of the limb and the lower part. It provides one dof in flexion and extension motion in sagittal plane but also allow slight degree of rotation of side-to-side movement.

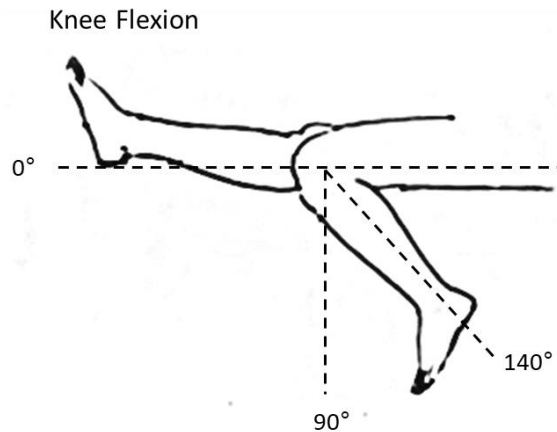

**Figure 8: Knee joint flexion and extension.**

- Flexion and extension in sagittal plane: Knee flexion is the movement that decreases the angle between the leg and the ankle whereas knee extension is the movement that decreases the angle between the leg and ankle. The flexion is about 140°, and the extension 0°[25]. As a gesture, these movements combine into a "kick" [26].

**2.1.1.3 Hip joint**

The hip joint is a "ball and socket joint" [25]. It allows flexion/ extension, abduction/adduction and outward/inward rotation. The movement of hip joint lead to motion of whole leg which introduce more fatigue. They often assist the movement of other joints. For example, abduction and adduction are used when moving foot mice or for free gestures horizontally [27].

### 2.1.1.4 Toes

Toe Joints are limited spheroid Joints, capable of sideways movement within certain limits, but are primarily intended as hinge joints with movement upwards and downwards. Alderson tested the toe-actuated electric arm with amputees, but abandoned them because they "proved difficult to operate with consistent precision" [28]. Because the toes are harder to control than the fingers and are often covered by socks or shoes, they are very seldom used for interaction. The exceptions are toe switches embedded into shoes, such as the one described by Carrozza et al. for the ACHILLE insole, that contains a toe switch for controlling a prosthetic arm [29].

### 2.1.2 Pose

Three main poses for foot interface are sitting, standing and walking/running. The foot interface is used for surgeon to control device in robotic surgery, thus, the situation of walking/running would be ignored. The focused poses are standing or sitting.

- Standing pose normally only allow one-foot operation, as the other foot should support and maintain the body's balance. Long time and complex task in standing pose cannot be realized because of problem of fatigue and body balance. To prevent fatigue, such complex gestures should be limited in time and motion space.
- Sitting pose provide stable and balanced position for one-foot operation and allows two feet lifting at same time [30]. However, for sitting position, a user's interaction range is limited to the shin and feet's reach. How the user was seated and the constrain of chair and desk may also affect the motion range and performance [27].

### 2.1.3 Foot capability

The feet are used as the primary modality for input when the user's hand are busy [22][31] They provide a natural mapping to locomotion tasks [32].Simeone et al. [30]conducted a user study and found foot-movements are easy to learn and can be used to perform 3D manipulations. However, like the other body gesture, the foot operation also has limits. The followings are some foot capability study found in the literature.

- **Accuracy:** Human lower limb can provide abundant information on movements though joints of hip, knee, ankle and toes. However, studies comparing the leg and arm show that it is not as precise as human hand [33]. The feet can be from 1.6 to 2 times as slow as the hands, but this difference can be reduced with practice. T. Han et al. [34] have investigated the foot kick gestures and how well users were able to control the direction of their kicks. They found that users have a best perform of their kicks when the movement range is divided into segments of at least 24°. And that users have difficulty in controlling the velocity of the kick, but can remember two broad ranges of velocity. Because of the limitation of accuracy of foot, Raisamo and Pakkanen suggest that the feet should be limited to low fidelity tasks, in which accuracy is not crucial [33], such as mode selection [35], non-accurate spatial tasks and performing secondary tasks whilst the hands are busy. However, D. Podbielski et al.[36] have compared the hand- and foot- activated surgical tools in ophthalmic surgery and found hand and feet have similar performance.
- **Force:** The feet can transmit small forces, such as for the operation of switches, in nearly all directions. The activation force of switch is normally not beyond 10N in standing operating pose in operation room (OR) [37]. However, ACHILLE interface uses a larger force threshold for foot switch imbedded in a wearable insole which are 69N (toe switch), 79N (heel switch), 112N (lateral internal switch) and 96N (lateral external switch) for controlling a prosthetic hand. In addition, the force of foot may vary according to the location of the pedal and sitting posture. Normally, the location of the pedals is directly underneath the body, so that the body weight above them provides the reactive forces to the forces transmitted to the pedals. Placing the pedals more forward makes body weight less effective for generation of reaction force to the pedal effort.

- **Mapping:** The mapping between foot gesture and task to be executed should be accord with the human logic. J Alexander [31] et al. 's suggest that users intuitively choose the foot gestures which are logical mappings from commands that they are familiar with. For example, users apply foot clockwise movement to command of rotating right; foot left kicks maps to navigation left command etc. The simplest form of a pedal is a binary switch. It also allows for controlling continuous parameters, but requires careful design considerations on how the user's input can be mapped to control commands. Kim and Kaber [38] have found the users tend to prefer rate-based ($1^{st}$ order) control approach than position-based (0 order) control in foot-based methods.

- **Visibility and feedback :** Moreover, similar with the other gestural interaction, foot movements may be less accurate when there is no visibility of the limbs. However, the lack of visibility of the feet can be compensated by the user's proprioception. Therefore, even though users are not able to see their feet they still know where they are in relation to their body. Haptic feedback can improve surgeon's performance during intervention [39]. Recent studies [40][41] have shown improvement in performance and learning enabled by using compliant interfaces providing the operator both force and position sensing. A compliant interface can give rich displacement and force feedback that may give the operator more accurate perception of their control actions, thus enable them more skillful operation even there is no assistance of visual.

## 2.1.4 Ergonomic guidelines

The foot pedals in OR are found not intuitive to use. Researches have collect feedback of 284 surgeons. A little more than half of the respondents feel it is uncomfortable to use the foot pedal. For example, no visual control over the pedal and some foot postures cause fatigue [42]. Thus, there are some guidelines for the use of foot interface in ergonomic literature, although some guidelines are used for force measure not control, they are still referential for us:

- Dorsal-flexion to control the foot switch should not exceed 25°[43].
- Enduring dorsal-flexion of the foot should be avoided to prevent discomfort of the tibia [43].
- The foot pedal must be controlled without looking at it [34][41]
- Feedback has to be provided without looking at directly at the feet [44].
- The space to carry out the task needs to be minimal [44]
- During control of the foot pedal the foot pedal may not move [34][41].
- The chance of accidentally activating the wrong switch function must be minimal[34][41].
- The chance of losing contact with the foot pedal must be minimal[34][41].
- Minimal time and physical effort required from the operator [45].
- Low height of the device to reduce the operator's workload [45].
- Easy and quick to adjust according to different anthropometrical features; able to use on both sides of foot[45]..
- Can be assembled into a medical system as a module[45]..

## 2.1.5 Summary

According to the examination of the foot biomechanics and human factors of foot operation, with the consideration of the application in surgical operation room, some initial design parameters are determined to satisfy the foot anatomy and ergonomic guidelines:

(1) One-foot operation was selected rather than two feet. As moving two feet was more tiring than moving just one. And in OR, another foot may already have occupied, for example, control foot pedal of diathermy.

(2) The interface is initial designed for sitting pose which is more stable without balance problem and introduce less fatigue than standing pose.

(3) Four dof of foot movements in left/right (hip abduction/adduction), forward/backward (flexion/extension of the knee joint), dorsiflexion/plantarflexion of ankle and lateral/medial axial rotation of the foot are selected as foot input motions. Eversion and Inversion of ankle and toe movement are not used in initial design because of their limitation of movement and poor control capability. The mid-air gesture of lifting foot and stepping which is commonly used in foot button/switch operation is not involved. Instead, gesture of dragging to left/right and forward/backward are less tiring and identical to human logic when controlling spatial movements.

(4) To cause less fatigue, all selected foot motions should be kept in a small range within the allowed motion range of foot biomechanics.

## 2.2 State of the art in foot interfaces of robotic surgery

### 2.2.1 Force control/activation interface

The feet can transmit small forces, such as for the operation of switches. The most applications are using the foot motion in downward or down-and-forward directions to activate buttons placed on the planar platform. The foot forces can be used as a binary activation or as continuous input.

#### 2.2.1.1 Binary switch interface

The button type interface is the most common foot interface format, it is robust, simple and easy to learn. They normally consist multiple binary switches/buttons, in which only two states of on/off can be achieved by one switch. This is a good choice to achieve single and simple function. For example, the previously mentioned EndoAssist [46] and FreeHand [47] systems which using head-tracking to control the robot also equip with a simple foot switch to prevent unintended movements: movement is enabled only when the pedal is in pressing state. Similarly, the TISKA system (Trocar and Instrument Positioning System Karlsruhe) [48] from Karlsruhe Research Center has a pedal to lock and unlock the motion of the tool. In addition, there are foot pedals/switches to control a microscope in microsurgery [49], activate a tool in ophthalmic surgery [50], control drilling tool speed in dental procedure [51], or to control a prosthetic hand [52].

Figure 9 shows two versions of a foot pedal in the Da Vinci system [53]. Instead of directly controlling a robotic arm, those pedals are used for activating the camera/tools, or robotic arm swapping. For the Da Vinci S system launched in 2006 shown in Figure 9(a), where all the buttons are located on a planar platform. The Si system launched in 2009 has a modified arrangement of the multiple foot pedals, which are located on upper and lower levels or vertically located on the side wall.

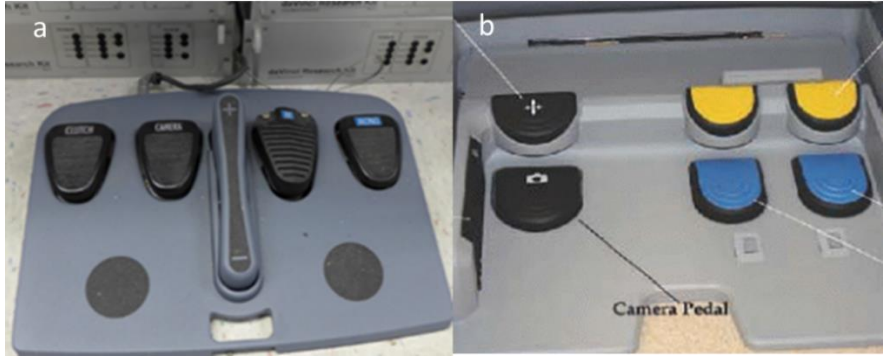

Figure 9: Foot panel of (a) DaVinci S foot panel (b) Da Vinci Si foot panel.

Except acting as auxiliary tools, there are some foot button interfaces available to control movements of endoscope in robotic surgery. Normally, one button is represented as one direction. When pressing down the button, there will be a constant speed for the slave tool. The speed may have several levels which can be selected beforehand. For example, the Endex endoscopic positioning system [54] of Andronic Device Ltd., features one single actuated dof. A two-treadle foot switch allows the operator controlling the scope to be inserted into or withdrawn out of the trocar. The Postural Mechatronic Assistance Solo Surgery (PMASS) [55] is a wearable endoscope holder presented in 2009 that allows navigation in three dof of movements, and up/down motion of the endoscope controlled by a foot switch.

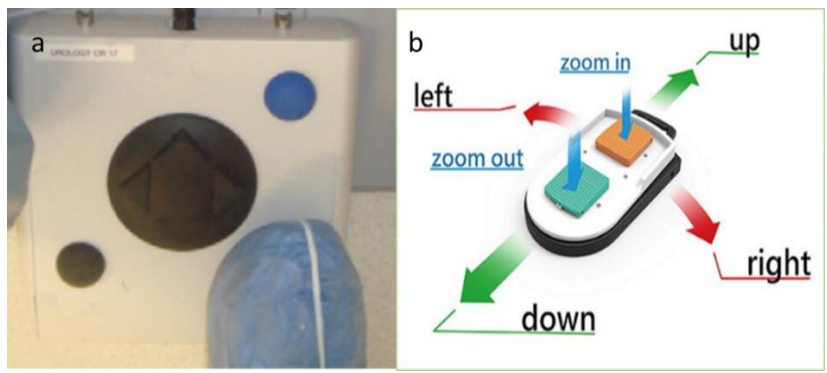

Figure 10:Foot interface of (a) ViKy (b) HIWIN.

USA patents 8591397B2 [56] and 2017/0303957A1 [57] describe a robotic endoscope holder system ViKy (Vision Kontrol endoscopY) by the Endocontrol Medical Company. As Figure 10(a) shows, this foot interface has multidirectional footswitch to control 3 dof of the robot [58]. The robot positioning corresponds to directions right/left, up/down or in/out movements of the endoscope could be controlled through surgeon's orders of foot. Another commercial robotic system is the MTGH100 endoscope holder [59] from HIWIN Technologies Corp, which features a six-direction control pedal. Surgeons can move the endoscope in six directions {in, out, up, down, left, right} via the foot pedal shown in Figure 5(b).

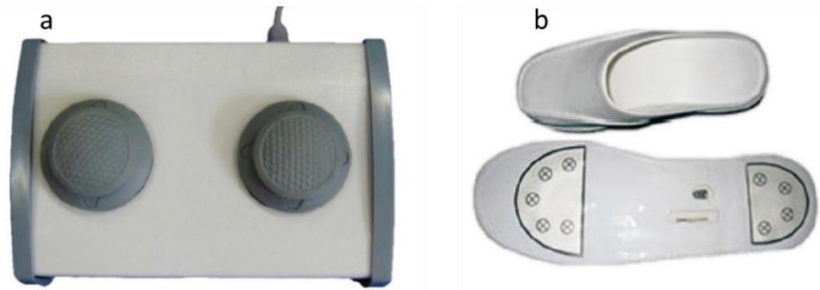

Figure 11: RoboLens foot interfaces (a) Foot switch (b) Wearable shoe.

The RoboLens [60][61] endoscope holder robot uses two kinds of foot interfaces. Firstly, it is equipped with a six-button footswitch which consists of two pedals (Figure 11(a)). The right pedal is used to control the lateral movement of the camera (up, down, left and right) while the pedal located on the left side allows the surgeon to control the zoom in and zoom out. In addition, a wearable foot interface with optical sensor can use both feet movements. It can control left, right, forward and backward directions using the right foot and forward and backward directions using the left foot. The comparison between existing commercialized interfaces and our designed interface is summarized in table A1 in Appendix in aspects of dof, control mode, force feedback, force measurement, foot adaption etc.

## 2.2.1.2 Continuous force control interface

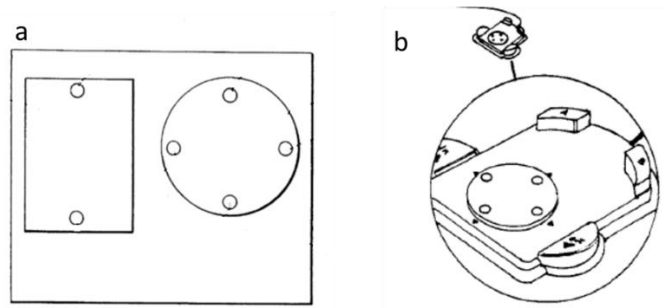

**Figure 12: (a) AESOP 1000 foot controller (b) AESOP 2000 foot controller.**

The foot force can act as continuous input to the system. For example, the surgical robot AESOP have a pressure-sensitive foot interface, the controlled velocity has a linear relationship with the foot force. As described in the corresponding USA patent [62] this foot interface has two switches with six transducers to control the endoscope movements (Figure 12 (a)). It can be used to move the endoscope in, out, left, right, up or down by applying pressure to the corresponding buttons on the foot controller that can be operated by the foot of a surgeon. In the system of AESOP 2000 [9] which mainly uses speech commands to control the camera, a foot controller as shown in Figure 4(b) can still act as an alternative interface to the voice interface. Compared to previous foot controller, the in/ out buttons are placed on the two sides of multiple directions buttons which can be controlled in a flexible way through the two feet.

Abdi et al. [18] have developed an elastic-isometric foot interface to control a robotic endoscope holder, Figure 13. This interface enables continuous control of three dof through pressing six pistons, and a fourth dof is provided by foot rotation around the vertical axis. However, the movements those interfaces could control are still limited to discrete directions, as it is not possible to command two or more dof simultaneously. Furthermore, the use of Hall effect proximity sensors to detect the spring force limits the control range thus for forces above 3N the slave robot could only move at a constant speed.

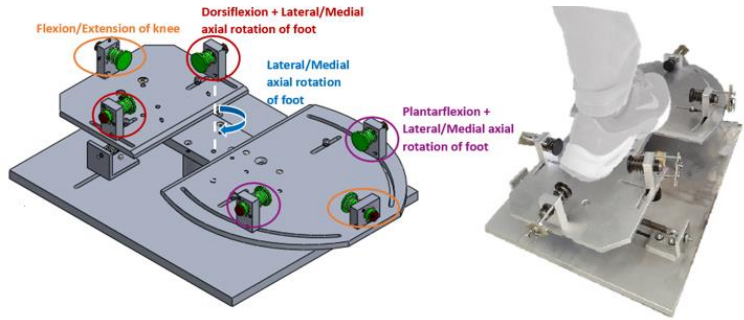

Figure 13: Elastic-isometric foot interface of Abdi et. Al.

### 2.2.1.3 The other force interface

There are some other types of foot interfaces for surgery developed using foot force information. For example, Kawai et al. used a pressure sensor sheet to record different foot patterns of the operator to locally control five dof including open and close of a surgical forceps [63]. However, the resulting signal propagation has a non-negligible delay as the sensor sheet cannot go back to un-deformed position immediately. Furthermore, the surgeon needs to memorize the meaning of patterns and change patterns by raising the foot and setting it down repeatedly, which will increase fatigue. Moreover, there is no haptic feedback provided to the surgeon.

### 2.2.2 Foot position tracking interface

Although the motion of foot is not as accurate as hand, the position information can still be used as a control commands. Chan et al. [64] collected foot rotations information through a IMU (inertial measurement unit) sensor which attached with foot using a strap to control movements of the endoscope. X Dai et al. [65] have similar design using IMU attached to operator's foot as shown in Figure 14. Although the position information of foot is collect, they are not used to map to position of robot but regarded as "position switch", for example, once the foot rotation angle beyond a threshold value, one joint is activated. Although the interface can control joints' movements of robot in four dof, each dof need to be activated separately and cannot be combined.

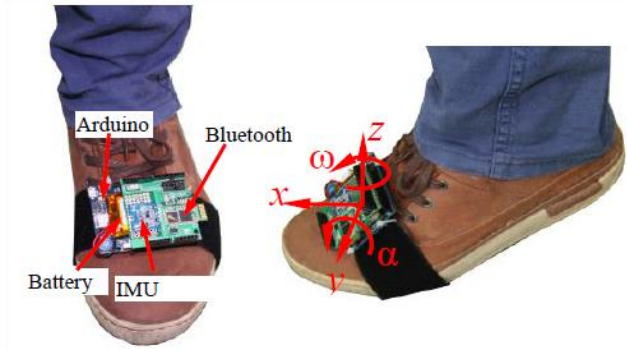

**Figure 14: IMU foot interface.**

Different from the IMU foot interface, Figure 15 shows an interface using position control [66]. A structure with two linear guides and two rotational joints are serial connected from the base to the pedal. Each joint was responsible for collecting one dof position information of foot. Specifically, the translational movements of the interface are tracked by magnetic membrane potentiometers. And rotational movements are tracked by encoders. Those position data will map to position of the robot. This control strategy has a mapping drawback of limited workspace of controlled device. The motion accuracy of slave device may not be guaranteed because of the foot motion error. In addition, the selected serial structure increases the total height of the foot interface and may cause coupling of foot motions. For example, the forward motion of foot may cause both movements of interface in direction of forward and pitch rotation.

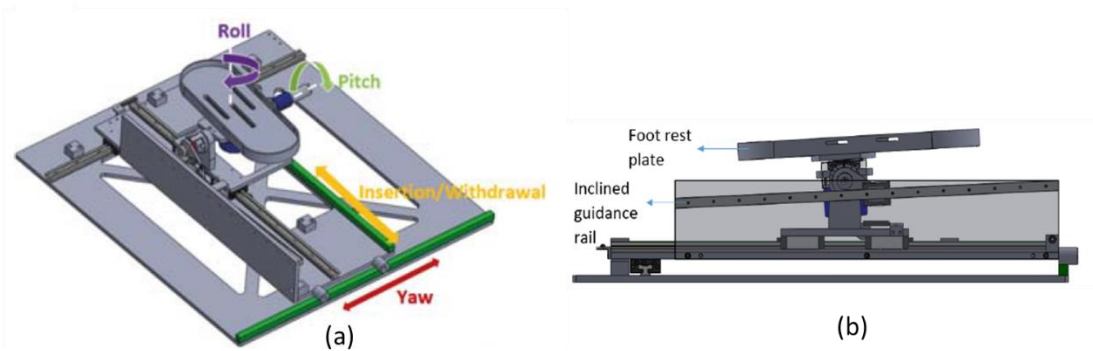

**Figure 15: Abdi's isotonic foot interface. (a) Prospective view. (b) Side view.**

Another example of foot interface collecting position information was shown in the Figure 16. Which is a 1 dof haptic foot pedal using position-rate control, the controlled tool's speed is proportional to the drift position of the pedal.

### 2.2.3 Haptic foot interface

Haptic feedback can improve surgeon's performance during intervention [39]. It can be separated into passive and active feedback according to whether the interface uses powered actuators(s).

The foot pedal shown in Figure 16 can control 1 dof of drilling tool's speed with active haptic feedback provided by a vibration motor [67]. The authors have proved that providing haptic feedback to foot can reduce the surgeon's response time to a warning signal. The active haptic feedback has the advantage to enable flexible adjustment of the resistance force/torque, however at the cost of a more complex control system.

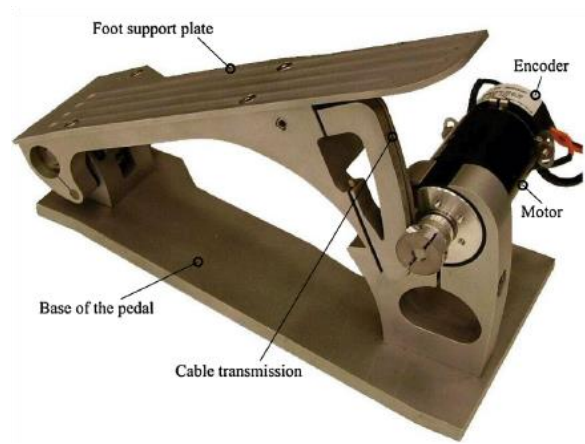

**Figure 16: Haptic foot interface.**

Passive haptic feedback is provided through the object's physical property [39], such as when pushing down a common pedal with passive spring offering resistance force during foot's movement. Passive haptic feedback mainly depends on the physical properties of the mechanical system, which can however be combined with visual feedback providing pseudo haptic feedback to the operator [68]. Although this feedback arises as haptic illusion [69], it is applied mechanically in the input device instead of electrically transmitted from a remote site, which avoids signal noise due to the feedback and greatly simplifies the system. Many experiments have tested the feasibility of pseudo-haptics in virtual reality [70][71][72]. For example, Kim et al. have tried to combine pseudo haptic feedback with a isometric hand manipulator [73] for robotic surgery by displaying force to the operator. However, we have not found any work using pseudo haptics on the foot channel.

### 2.2.4 Summary

Although many foot interfaces for robotic surgery have been developed, they are not sufficient to control a robotic arm efficiently and intuitively and in conjunction with the (natural) hands. The shortcomings of these interfaces are summarized below which illustrates the needs for both a suitable interface and appropriate signal processing.

(1) Existing foot interfaces are mainly composed of switches, pedals and buttons. The number of functions that can be controlled by these devices is determined by the number of buttons. For instance, to control the robotic arm with laparoscopic camera with 4 dof, at least eight buttons are needed. In that case, the surgeon is like playing acrobatics. The risk of pressing wrong button is increased. In addition, the mid-air foot gestures may bring more human fatigue which will affect the operator's performance and operation time.

(2) Degree of freedom(dof): The current commercial foot interfaces mentioned above e.g. AESOP, ViKy, RoboLens etc. could control at most three dof which is less than required to move the trocar in keyhole surgery.

(3) Control mode: Although some interfaces designed in academia increase the motion they can control to four dof or more, the control signals are mostly restricted to a few discrete foot-based input signals. Which means the movements they can control are no more than single (Cartesian) axes "forward", "backward", "left", "right", etc. without capability to control the movements in combined continuous directions. This is less efficient for controlling spatial movement, as any movement need to separate to sub-motion of multiple single Cartesian directions.

(4) The lack of haptic feedback may require the operator to visually check the foot posture frequently, making the control more complex and causing fatigue. And no multi-dof haptic foot interfaces were found during the current literature review.

Therefore, in the following chapters, a passive and compliant 4dof foot interface will be presented yielding force and position feedback, with built-in technique to extract operator specific commands based on natural foot movement patterns thus enabling intuitive and efficient control of a slave mechanism.

# Chapter 3 Design and development of the foot interface

## 3.1 Introduction

This chapter will explain systematically the designing process for the foot interface.

### 3.1.1 Design requirements/objective

From the overall project objective described at Section 1.3, and the reviews on critical biomechanical aspects of human foot (Section 2.1) and the application and current interfaces (Section 2.2), the following requirements of the interface:

(1) The interface should enable to control at least four dof movements of a slave robot; the multi-dof foot information should be collected without coupling.

(2) The motion it controls should not limited to single Cartesian directions, but any free spatial movements according with human logic.

(3) The interface should have positioning assistance mechanism to facilitate the control of the operator, ergonomically helping the surgeon position his/her foot without need of visual check.

(5) The interface should have a foot resting place which provide resting position when foot operation is in idle state and this position should also be convenient to start new operation without reposition the foot. However, the risk of accidental activation should be avoided when "riding the pedal" [74].

(6) The device should have a foot adjustment mechanism to fit different operator's foot size or/and preferred posture positions. As the dominant leg vary from operators, the interface should also be able to use on both side.

(7) It is preferable to have a subject-specific mapping from user's intention to control command as the parameters of foot gestures vary from person to person. And the control performance should also not be affected a lot by the sitting chair, operation desk and habitual sitting postures.

(8) It is preferable if the interface allow coarse manipulation and fine manipulation of slave robot.

**3.1.2 Design specifications**

To satisfy the above requirements, a novel interface will be presented with the following features:

(1) The developed interface uses parallel-serial hybrid structure which effectively achieve four dof control without coupling of motion. In addition, the movement is not limited to a single discrete direction but in contrast can yield any combination of directions in four dof. For instance, the robotic arm can be controlled by the foot to move in diagonal directions (e.g. both forward and right) and modify the direction continuously.

(2) The use of eight springs (six compression springs and two torsion springs) forms a compliant system which can provide a feedback in force and position via the elastic interaction between the interface and the operator's foot. The movement distance of the pedal assembly is increasing monotonically with the imposed force, through which the surgeon could sense the pedal's position and thus receive real-time dynamic haptic feedback. This allows the surgeon to control various conditions without having to control the foot position using vision, thus improving efficiency and reactivity.

(3) The interaction force information between the foot and the interface is collected through eight load cells coupled one end of one of the eight springs and receiving the corresponding elastic force. When some spring is fully compressed, the corresponding load cell can still record the increasing force yielding a continuous transition between the elastic input and isometric input modes. The larger force control signals above the spring full compression force will still be recorded, which can be used for control.

(4) The interface has an automatic positioning mechanism comprising multiple springs yielding a single minimum of elastic energy. This can bring the system to home in the posture corresponding to this global minimum when the foot is relaxing, which is the position of the interface with pedal assembly located in the center of the base space. Once the surgeon finishes operation movements and releases the pedal, the interface will go back to this neutral position automatically, thus providing a resting position for the operator's foot.

(5) The force information can be used to provide visual feedback of an interaction of e.g. a robotic arm with a dynamic environment, thus enabling control of the interaction with this environment in position and force with visual and (pseudo-) haptic feedback.

(6) The interface software has a built-in technique and algorithm to extract operator specific commands based on natural foot movement patterns thus enabling intuitive and efficient control of a slave mechanism e.g. a robot arm equipped with surgical tools.

## 3.2 Mechanical design

The foot interface will be used to control the locomotion of a slave robotic device. A closer match between control and natural actions would provide better motor substitution, making it more natural for the operator to perform and resulting in a more realistic simulation. Thus, several specific foot motions are selected as input to the foot interface system as Figure 17 shows, which are forward backward movements (resulting from the knee joint flexion/extension), lateral movements (from the hip's abduction/adduction), dorsiflexion/plantar flexion of the ankle and foot's lateral/medial axial rotation. These four movements can be used to activate pedal movements in defined $x$ and $y$ axes, pitch rotation and yaw rotation respectively with respect to the base.

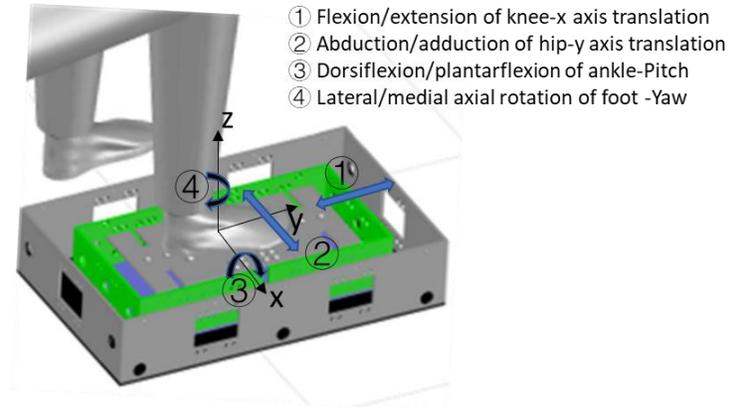

**Figure 17: Four dof of foot interface and related human movements.**

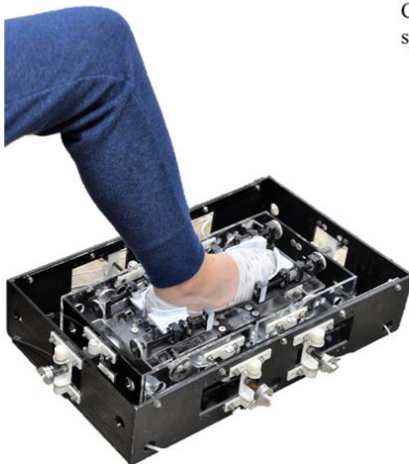
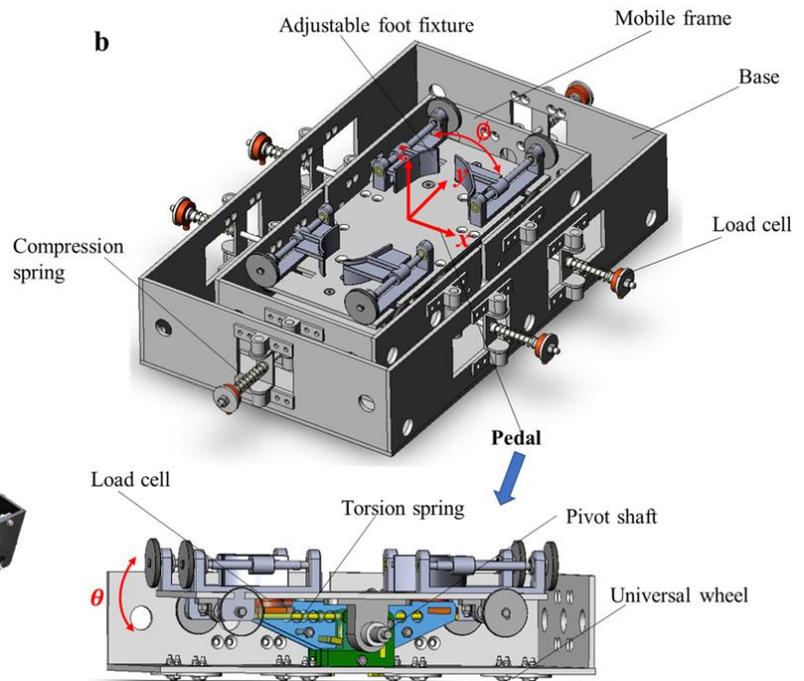

**Figure 18: Passive 4dof foot interface. (a) photo of the interface prototype with a user's foot. (b) 3D system mechanical structure in perspective view and side cross view.**

As shown in Figure 18 (more figures are listed in the Appendix A1), this interface is composed of: 1) a flexible movable pedal assembly (Figure A4 and A5), 2) a base assembly (Figure A3), 3) eight springs (six compression and two torsion springs) together with eight force sensors (Figure A6 and A7). The pedal assembly includes a mobile frame and a treadle portion. The treadle and the mobile frame are connected through a shaft, so that the treadle plate has one dof in the direction of the ankle dorsiflexor-plantar-flexor movement of the human ankle. The treadle is supported by the shaft and one pair of torsion springs with one end is fixed on the mobile frame and the other to a load cell. The pedal assembly is sliding on the surface of the base and can move freely in $x$-$y$ plane. To simplify the mechanism, no revolute joint or linear guideway is used to control the three dof of $x$ and $y$ translations and yaw rotation. Eight universal wheels are mounted at the bottom of the mobile frame to minimize friction. The pedal assembly is connected to the base through six linking guides with hinge joints on both sides, with a compression spring and a donut compression load cell on each linking guide to measure the forces applied by the operator's foot.

### 3.3 Spring mechanism

### 3.3.1 Selection of springs

The springs are key components of the foot interface, as they provide haptic feedback to the surgeon and define the home position. The $x$, $y$ and yaw dof, with refer to Figure 17, are designed using the same six springs. Common spring of compression spring and tension spring are two choices. The comparison between compression springs and extension springs are listed in the following Table 1. For our application, compression springs are more suitable than extension springs, because it is easy to assemble force sensor and enables combination of elastic and isometric control mode.

Table 2: Comparison between compression springs and extension springs.

| Comparison aspects | Compression springs | Extension springs |
|---|---|---|
| Structure | Need guides | No need of guides |
| Force sensing | Easy to assemble force sensor | Need extra mechanism to support force sensor |
| Control strategy | Easy to combine elastic and isometric control modes, as compression spring become rigid when fully compressed. | Difficulty to implement the transition between the elastic and isometric modes; need to figure out how to sense force continuously after a spring is fully stretched. |

The pitch dof of pedal is separated from the above three dof. A pair of torsion springs are chosen as they can be installed in a simple system and yield a linear relation between the pitch rotation angle and the torsion spring force. The six compression springs have the same stiffness of 200N/m. Springs of stiffness {100, 200, 300, 500, 1000} N/m were tried, however the 200N/m stiffness springs were selected by trial-and-error, as they were not too hard to press while providing sufficient haptic feedback to control force.

### 3.3.2 Home position with minimum of global elastic energy

Instead of locating spring between the mobile frame and the base frame, six compression springs are designedly placed outside the base frame. Figure 19 (a) shows the former structure in schematic diagram. This design has a poor self-center mechanism. The compression spring system will only stable in the position with minimum global elastic energy. The resultant torque making the pedal rotate is much bigger than the torque helping it go back to home position. The extreme rotation positions shown in Figure 19 (b) and (c) have smallest spring potential energy than the home position which cause singularity in the workspace. Thus, after the external force removed, the pedal assembly tends to stay in the extreme rotation positions rather than going back to home position.

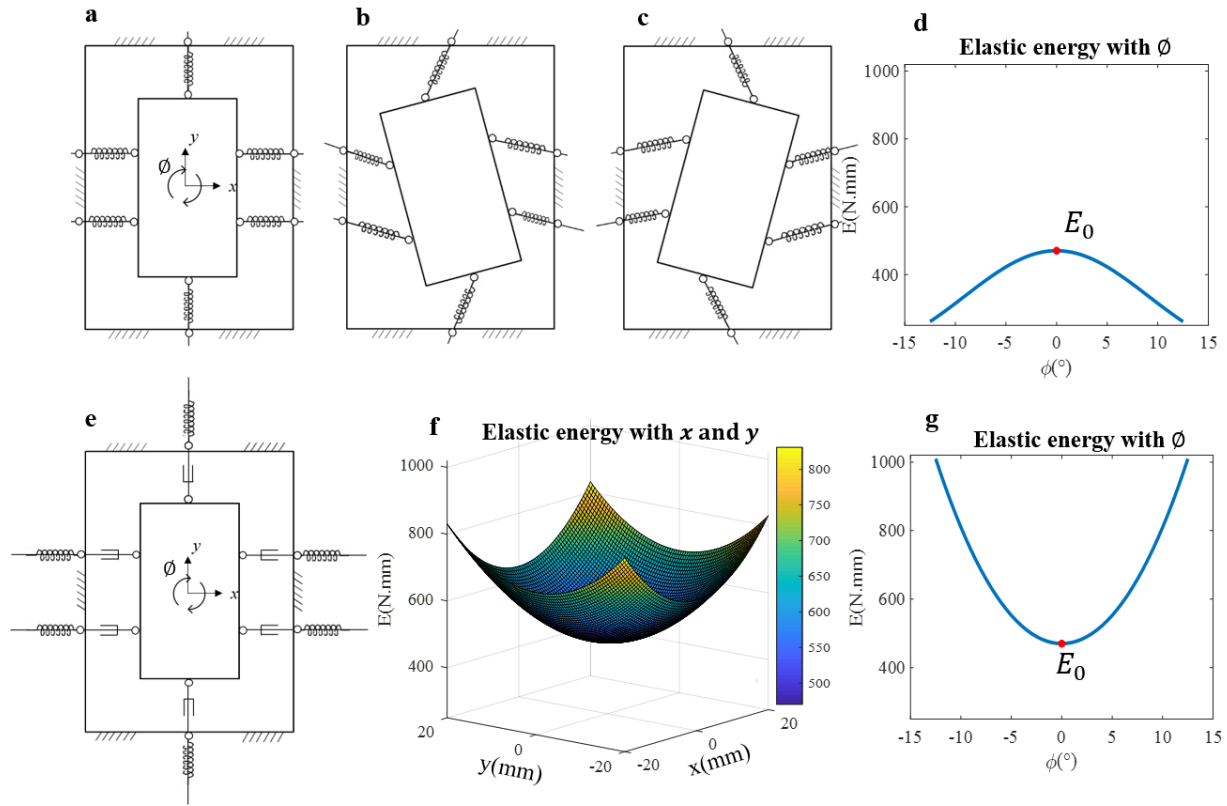

**Figure 19: (a)Initial design with springs between the base and MF (b,c) exhibiting two singular postures and (d)elastic energy changing curve with yaw rotation angle.(e) Improved design with (f,d,g) global energy minimum.**

In comparison, putting the compression springs outside the frame define a well home position. In which, the two torsion springs are balanced and support the treadle plate horizontally; Six compression springs are equally compressed with pre-tension forces of 5.6N and 2.8cm deflection (For each spring, the maximum deflection is 4.8cm and free length is 6.0cm), keeping the pedal in the middle of the base. The resultant force and torque are zero and a minimum of global elastic energy for the whole system is achieved at the home position. Once the operator releases the pedal, the interface will return to that home position automatically, providing a chance for the operator to rest his or her foot.

### 3.3.3 Haptic feedback

The eight springs form a compliant system providing haptic feedback of force and distance via the interaction of the operator's foot and the pedal. In the foot interface, the haptic feedback is provided in a dynamic and real-time way in accordance with the operator's foot motions. The 3 dof in two translations and one rotation inside the base horizontal plane will lead to movements of six spring guides. The movement of each guide is coupled with the varying force of the responding compression spring. Different pedal movements will bring different compressing states for the springs' system. The combined motions of any single directions will bring combined haptic feedback. For instance, the operator will receive diagonal displacement and force feedback when they conduct diagonal motion. The haptic feedback related to the teleoperation environment will be tested and studied in the future as pseudo haptic feedback.

### 3.4 Transmission between elastic mode and isometric mode

Our interface combines elastic and isometric input modes and provides a force range in all four dof limited only by the sensor properties. The forces/torque exerted by the operator's foot are the inputs to the interface. Generally, the exerted force can be measured from the spring compression or from its deflection. In our design, we choose to measure the compression using an affordable load cell placed in series to this spring, which has the advantage to avoid a ceiling effect when the spring is fully compressed, and to continue measuring forces above the full compression force. Resultant force and torque on the pedal could be computed from the forces applied on the 8 load cells. For example, when the pedal moves forward, the spring located at the back side will deform first till fully compressed, if the user further pushes forward, the force sensor placed at the back side will continue to record the force, which is then used to control the slave robot arm.

## 3.5 Adjustable foot fixture mechanism

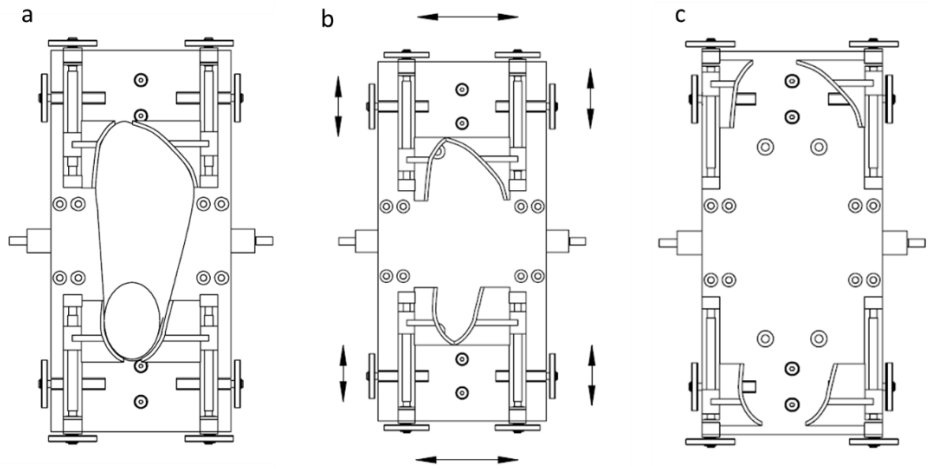

**Figure 20: Adjustable foot fastening mechanism. (a) Illustration of fitting a foot (b) Minimum adjustment size. (c) Maximum adjustment size.**

An adjustable foot fixture mechanism attached to the pedal enables an operator specific adjustment to the foot size yielding comfortable but rigid fixation thus efficient operation for an extended time. It is composed of four foot-shaped guides connected to eight guide screws to achieve flexible adjustments in length and width by easily twisting the handle. The detailed drawing can be found in Appendix Figure A1.5. The foot-shaped guides (fabricated using 3D printing) can fit well different human foot anthropometry (Figure 20) from foot size 35 to 46 (EUR standard) [75].

## 3.6 Workspace analysis

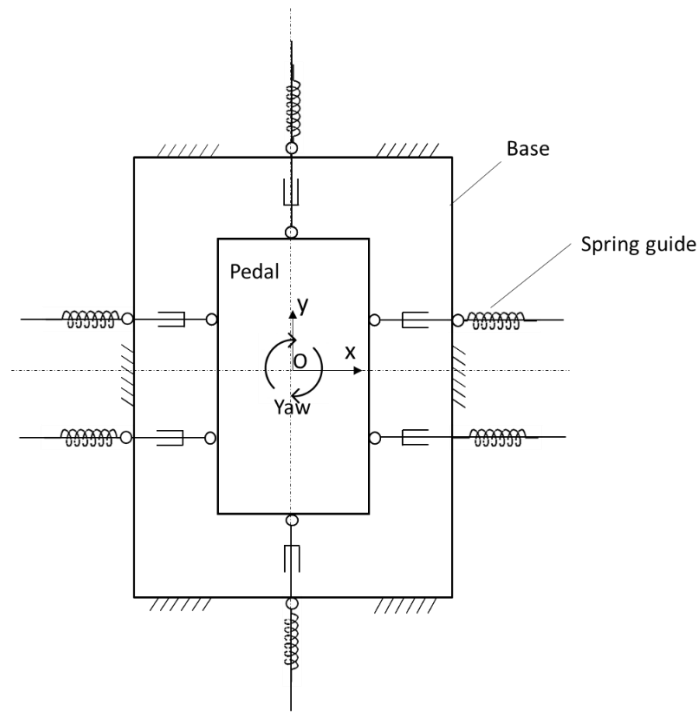

**Figure 21: Schematic diagram of the foot interface.**

Figure 21 shows a top view diagram of the foot interface. The six spring guides are connecting the base and the pedal, which form a planar parallel structure[76]. The pedal in the middle could move freely in the spacing of the base in two translational dof and one yaw rotation on the horizontal plane. The workspace of the translations in *x* and *y*, and how the workspace changes with the yaw rotation angle was analyzed through using a Monte Carlo method[77].

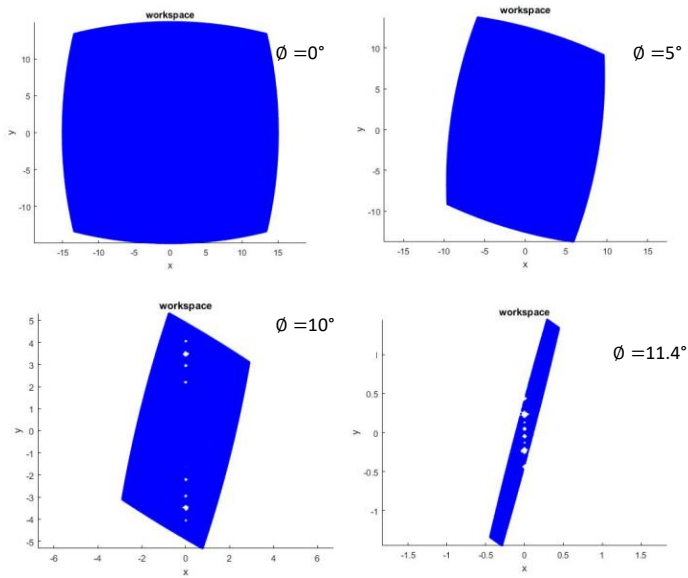

**Figure 22: Monte Carlo simulation and workspace changing with yaw rotation angle. (a) Translation workspace of x and y when rotation angle ∅ =0; (b) Translation workspace when ∅ = 5°; (c) Translation workspace when ∅= 10°; (d) Translation workspace when ∅= 11.4.**

Figure 22 shows how the workspace depends on the yaw rotation angle. When there is no rotation in yaw, the pedal could move freely in any single or combined positions in this an area as Figure 22 (a) shows. The translation workspace will decrease with the yaw rotation angle increase. Once the yaw rotation angle reaches the maximum value, the translation in x and y will have no space to move. However, within the extreme positions, displacement and rotation in each DOF could be manipulated together.

The pitch DOF is independent from the other three dof. Its moving range was set mechanically through a stop block. The motion range of the interface is described in Table 3, exhibiting values encompassing the typical human range.

**Table 3: Motion ranges of the foot interface.**

|  | x | y | Yaw rotation | Pitch ration |
|---|---|---|---|---|
| Motion Range | ±20mm | ±20mm | ±12.5° | ±10° |

## 3.7 Kinematics and statics analysis

### 3.7.1 Parallel structure kinematics analysis

The resultant forces on the foot treadle can be calculated through statics modelling based on the structure's geometry. To analyze these statics, it is assumed that the base frame, foot pedal assembly and spring guides are in the same plane. Therefore, the kinematics model of the foot interface (except pitch rotation) can be regarded as a 6-RPR planar parallel mechanism [78] without active joints . The mobile pedal is constrained by six spring guides. The attachment point on the base side is noted $A_i$ while the corresponding attachment point on the pedal side is noted $B_i$.

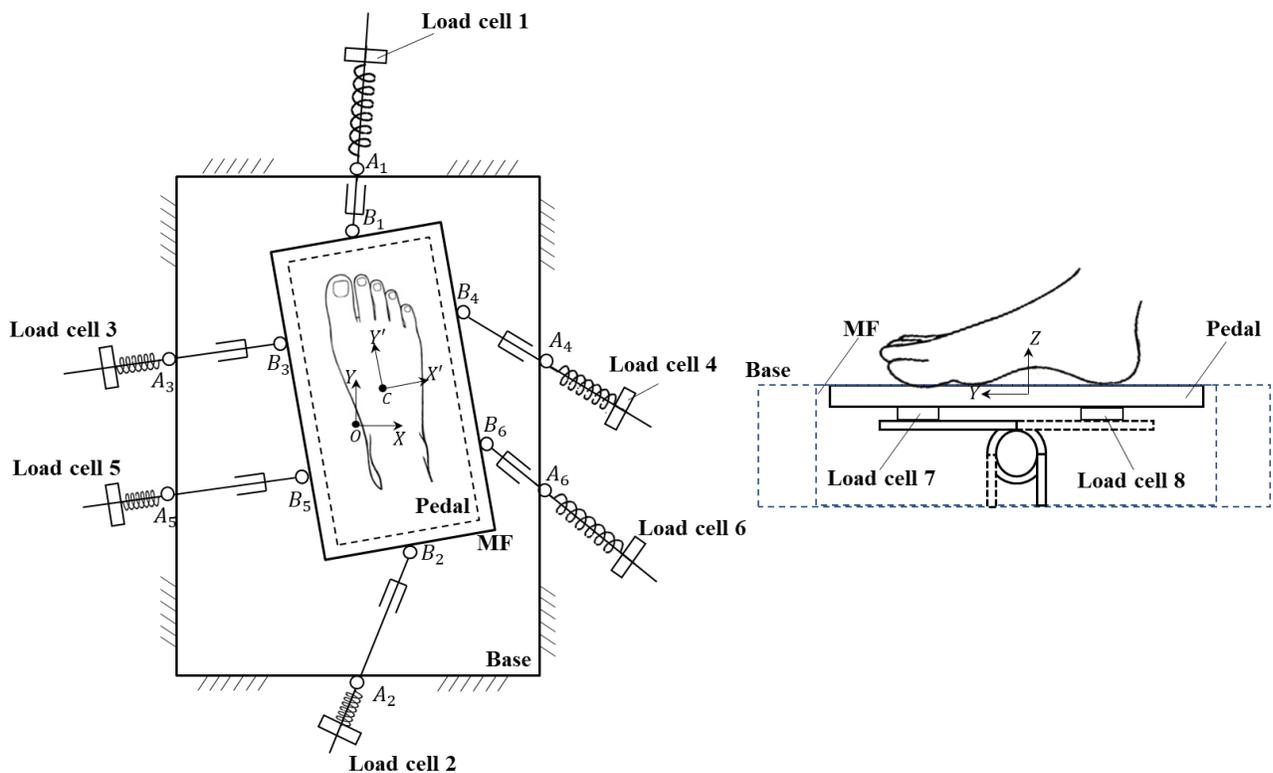

**Figure 23: Kinematics model of the FI.**

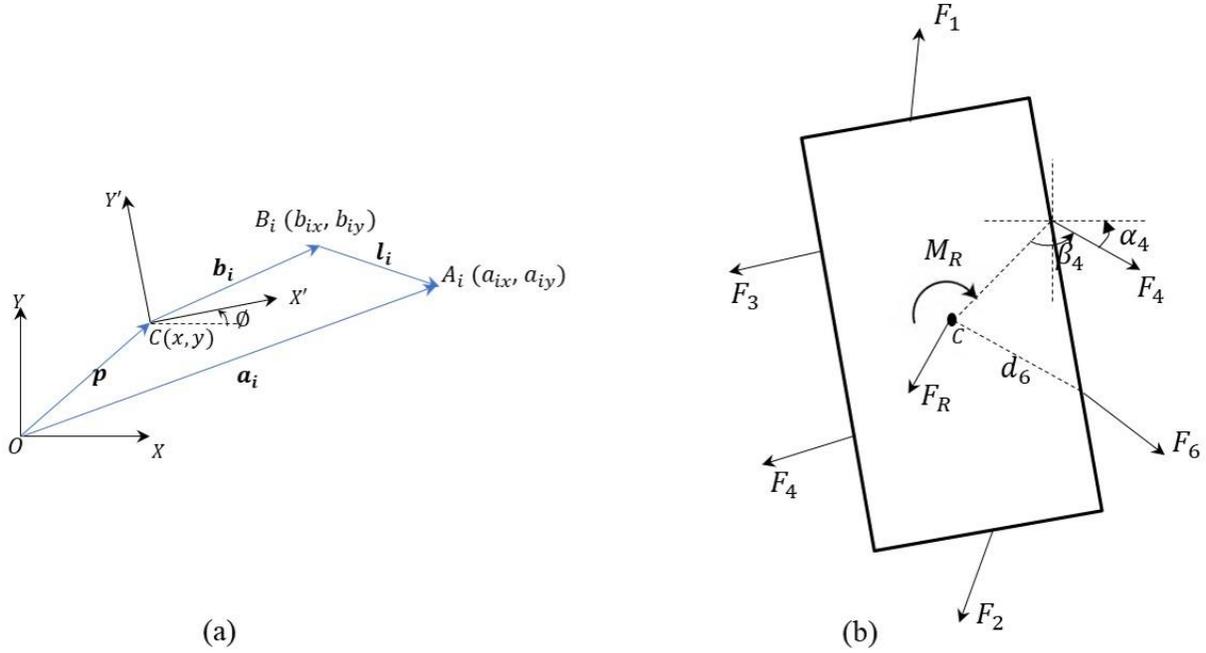

**Figure 24:(a) Kinematic model of *i*th spring. (b) Force analysis in x-y plane for mobile frame.**

Each compressing spring forms a kinematic chain. As shown in Figure 24, a fixed frame {O} and mobile frame {C} are assigned to center of the base and the pedal respectively. Let $p$ be the position vector of the pedal center point C expressed in fixed coordinate {O} and $\emptyset$ be the pedal rotation angle from the $\widehat{x_O}$ axis of the {O} frame to the $\widehat{x_C}$ axis of the {C} frame. The length of each spring guide between the base and the pedal is denoted as $l_i$; $a_i, b_i, i = 1, \ldots, 6$ are further defined as shown in Figure 25. From these definitions, a vector loop-closure equation [79] can be written for each spring:

$$\boldsymbol{l_i} = \boldsymbol{a_i} - {}^C_O R \boldsymbol{b_i} - \boldsymbol{p} \qquad (1)$$

where ${}^C_O R$ is the rotation matrix from the fixed frame {O} to the mobile frame {C}, written as

$${}^C_O R = \begin{bmatrix} \cos\emptyset & -\sin\emptyset \\ \sin\emptyset & \cos\emptyset \end{bmatrix}. \quad \boldsymbol{p} = \begin{bmatrix} x \\ y \end{bmatrix}, \quad \overrightarrow{OA_i} = \boldsymbol{a_i} = \begin{bmatrix} a_{ix} \\ a_{iy} \end{bmatrix}, \quad \overrightarrow{CB_i} = \boldsymbol{b_i} = \begin{bmatrix} b_{ix} \\ b_{iy} \end{bmatrix}.$$

Given the spring length, Equation (1) is written for each spring, each equation is squared to yield scalar equations (2) in the three unknowns $X = [x \quad y \quad \emptyset]^T$:

$$x^2 + y^2 + a_{ix}^2 + a_{iy}^2 + b_{ix}^2 + b_{iy}^2 + 2b_{ix}\left((x - a_{ix})\cos\emptyset + (y - a_{iy})\sin\emptyset\right) +$$

$$2b_{iy}\left((y - a_{iy})\cos\emptyset - (x - a_{ix})\sin\emptyset\right) - 2(xa_{ix} + ya_{iy}) - l_i^2 = 0 \qquad (2)$$

Then, the solutions could be represented as:

$$\emptyset = \arcsin\left(\frac{E}{2P}\right)$$

$$x = \frac{F(a'\sqrt{4P^2 - E^2} - 2aP) - 2bEG}{4Q\sqrt{4P^2 - E^2} - 8PM}$$

$$y = \frac{2G(b'\sqrt{4P^2 - E^2} - 2bP) - a'EF}{4Q\sqrt{4P^2 - E^2} - 8PM}$$

If let $E = l_4^2 + l_5^2 - l_3^2 - l_6^2$; $F = l_3^2 + l_5^2 - l_4^2 - l_6^2$; $G = l_2^2 - l_1^2$; $M = ab + a'b'$; $Q = ab' + a'b$; $P = bc' - b'c$

$l_i$ can be derived from Hooke's law and the load cell readings.

$$l_i = (F_i - F_{0i})k_i^{-1} + l_{0i} \qquad (3)$$

Where $F_{0i}$ is the pre-tension force of $i$th spring and $F_i$ is the responding vector force of $i$th spring, which will be reflected by the load cell readings. $k_i$ is the compression spring stiffness constant; $l_{0i}$ is the initial length of spring guide at zero state. In elastic mode, the directions of $F_i$ noted as $\alpha_i$ (Figure 25 (b)) can also be derived through the geometric relationship:

$$\overrightarrow{A_iB_i} = \overrightarrow{OB_i} - \overrightarrow{OA_i} = \begin{bmatrix} x + \cos\emptyset b_{ix} - \sin\emptyset b_{iy} - a_{ix} \\ y + \sin\emptyset b_{ix} + \cos\emptyset b_{iy} - a_{iy} \end{bmatrix} = \begin{bmatrix} x_{AB} \\ y_{AB} \end{bmatrix}$$

$$\tan\alpha_i = \frac{y_{AB}}{x_{AB}} = \frac{y + \sin\emptyset b_{ix} + \cos\emptyset b_{iy} - a_{iy}}{x + \cos\emptyset b_{ix} - \sin\emptyset b_{iy} - a_{ix}}$$

$\beta_i$ is the angle between vector $\overrightarrow{CB_i}$ and $\overrightarrow{A_iB_i}$, which represent the angle between the force vector $\boldsymbol{F_i}$ and the lever arm vector $\boldsymbol{b_i}$:

$$\cos\beta_i = \frac{\overrightarrow{CB_i} \cdot \overrightarrow{A_iB_i}}{|\overrightarrow{CB_i}||\overrightarrow{A_iB_i}|}$$

When regarding the foot interface structure as a rigid, the resultant force at point C in *x* and *y* directions noted as $F_x$ and $F_y$ can be obtained:

$F_{ix} = F_i \cos\alpha_i$, $F_{ix}$ is *x*-direction component force for $F_x$

$F_{iy} = F_i \sin\alpha_i$, $F_{iy}$ is *y*-direction component force for $F_y$

$$F_x = \sum_{i=1}^{6} F_{ix}$$

$$F_y = \sum_{i=1}^{6} F_{iy}$$

$\boldsymbol{M_i} = \boldsymbol{F_i} \times \boldsymbol{b_i} = |\boldsymbol{F_i}| \cdot |\boldsymbol{CB_i}| \cdot \sin\beta_i$ is the moment due to *i*th spring compression force around mobile frame mass center C. The total torque can be represented as $\boldsymbol{M}$ (define anti-clockwise as positive direction (+) and clockwise as negative direction (–) wrt. point C).

$$\boldsymbol{M} = \sum_{i=1}^{6} \boldsymbol{M_i}$$

It is assumed that the external forces from human foot $\boldsymbol{F_F}$ and moments $\boldsymbol{M_F}$ acting on the pedal are equal to the resultant force $\boldsymbol{F}$ (resultant force of $F_x$ and $F_y$) and torque $\boldsymbol{M}$ applied by six springs to the mobile frame. As such, the statics equations are:

$$\boldsymbol{F} = \boldsymbol{F_F}$$

$$\boldsymbol{M} = \boldsymbol{M_F}$$

## 3.7.2 Isometric state statics and the fourth dof

In the elastic state, the movements are allowed and the force and displacement have a linear relationship. The translation movement ranges of geometric center of the pedal are shown in the Section 3.6 workspace analysis. Once the pedal reaches the limits of the workspace, it cannot move anymore and transmit from elastic state to isometric state. The values of variables like $x$, $y$, $\emptyset$, $\alpha_i$ and $\beta_i$ which are related to geometric parameters will not change. Whereas if the operator exerts larger forces, the angle of force $\alpha_i$ will keep the values, and the corresponding load cell can still record the magnitude of increasing force. The resultant force and torque are calculated in same method using $F = \sum_{i=1}^{6} F_i$ and $M = \sum_{i=1}^{6} M_i$. For example, as Figure 18 shows, the pedal moves forward leading to compression and release of springs. It reaches the extreme position when spring 2 is fully compressed and becomes rigid. After that, if the operator continues pushing forward, the load cell 2 will continue to record the isometric forces. This design will avoid the ceiling effect of control signal.

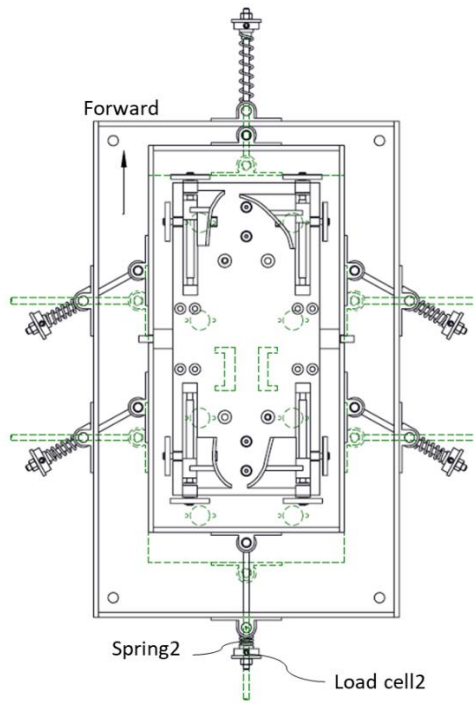

**Figure 25: Isometric state in forward motion extreme position**

The fourth dof of pitch rotation is activated by toe up and toe down rotation of foot (Figure 24) with maximum 10° rotation angle. The force information will be recorded by two load cells below sole and heel represented as $F_7$ and $F_8$. As the pitch rotation angle is small, the $F_7$ and $F_8$. Are approximate to component forces in vertical direction. Thus, we can move $F_7$ and $F_8$ to the center of the pedal and define $F_7$ in directions $Fz$ (+) and $F_8$ in $Fz$ (-) respectively.

## 3.8 Electronics and communication

The communication within the interface system is described in Figure 26. The foot interface is acting as a master console directly controlled by the operator's force. The load LW1025-25 from Interface, Inc., USA is used which can measure up to 25lbs. The load cells signals are amplified by 91 to 2222 times, then transmitted through a microprocessor to the computer using serial port. The chosen main board is the Arduino Mega 2560 with operation voltage to be 5V and 16 analog input pins. The computer can be connected with a slave robot e.g. via TCP/IP.

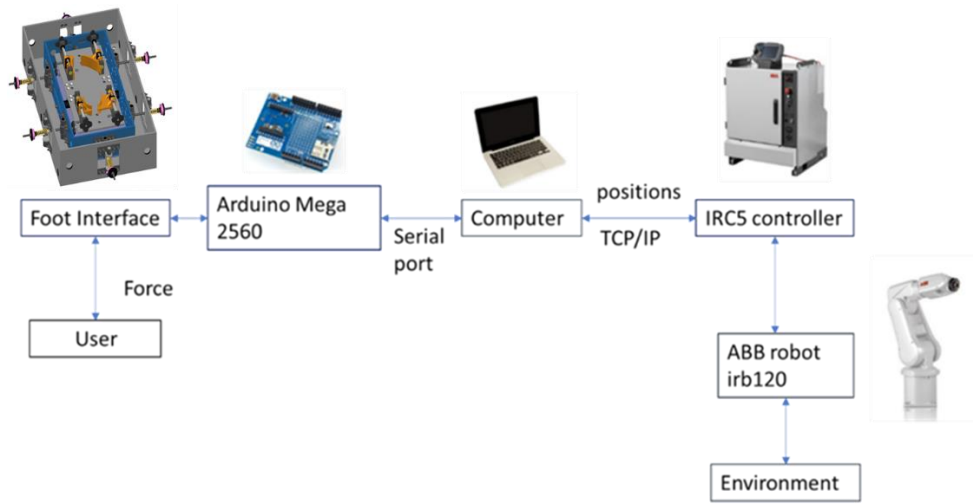

**Figure 26: Communication of the tele-operation system controlled via the foot interface.**

# Chapter 4 Intuitive mapping methods analysis

To make the interface best reflecting the operator's specific movement parameters, experiments (Section 4.1) were carried out to collect human foot movement information using the foot interface. Various mapping and their capability to catch desired foot movements in Cartesian directions were investigated through metric of prediction accuracy of directions (Section 4.2). The Statics modeling method and Independent Component Analysis (ICA) method are tested in Section 4.3 and Section 4.4 respectively. The other identification methods like k-nearest neighbor's algorithm (kNN) are also tested and the result is shown in Appendix A2.

## 4.1 Foot movements data collection

The experimental trials were conducted with 10 voluntary subjects (age 23-30, four females) to collect the motion data. The experiment protocol was approved by the Institutional Review Board (IRB) (Reference: 2018-05-51) on 3$^{rd}$ July, 2018, and the methods were carried out in accordance with the approved guidelines. Two types of motion were conducted in directions of single Cartesian axes, and diagonally, through the combination of two Cartesians axes. The force information of human foot will be collected through eight load cells of the foot interface at 50Hz data collection rate.

## 4.1.1 Single directions motion test

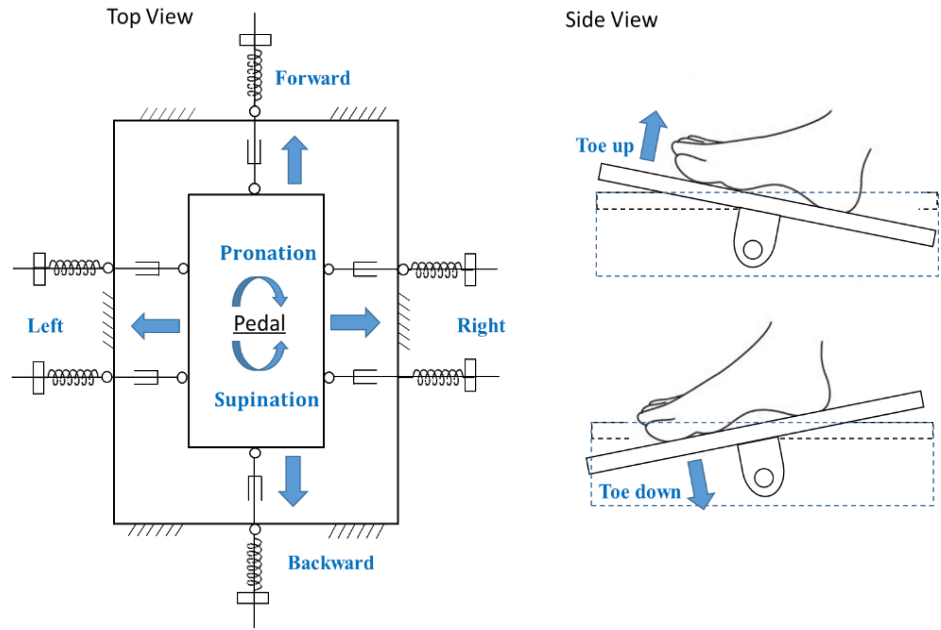

**Figure 27: Eight single axis motion of the foot pedal.**

Subjects were requested to move the pedal from the home position to the extreme position in eight directions {left(L), right(R), forward(F), backward(B), toe up(TU), toe down(TD), supination(S), pronation(P)} as illustrated in Figure 27, and then go back to home position. Three trials were carried out in each direction where the extreme position had to be kept for 1 second.

## 4.1.2 Diagonal directions motion test

j

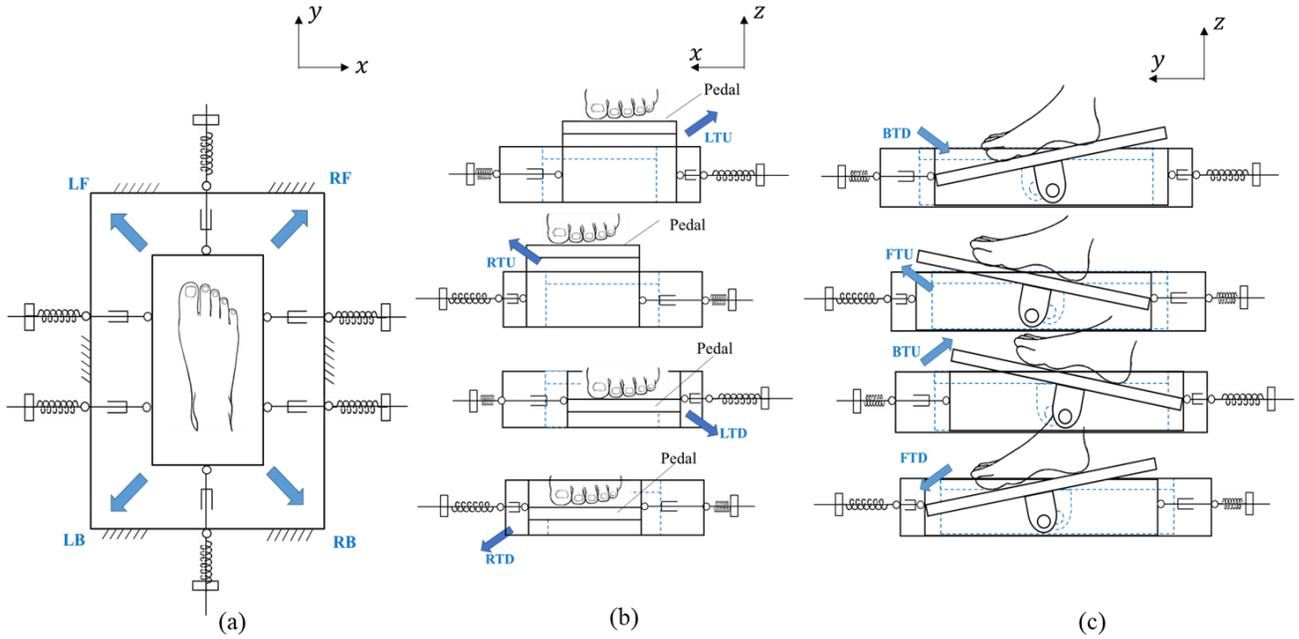

**Figure 28: Twelve diagonal motions of the foot pedal.**

In the second experiment, the subjects were instructed to move the pedal to twelve combined directions {left & forward (LF), right & forward (RF), left & backward (LB), right and backward (RB), left & toe up (LTU), right & toe up (RTU), left & toe down (LTD), right & toe down (RTD), forward & toe up (FTU), backward & toe up (BTU), forward & toe down (FTD), backward & toe down (BTD)} which are shown in Figure 28. Three movements were carried out in each direction where the extreme position had to be kept for 1 second.

## 4.2 Motion mapping criterion

As Figure 29 shows, the forces estimated by the 8 load cells {$F1, F2…F8$} of the foot interface are mapped into motor commands {$Fx, Fy, Fz, M$} to control the robot's movements. In each time stamp, there is a related force vector $\boldsymbol{F} = [Fx, Fy, Fz, M]$ yield out. The directions of foot movements can be represented under the coordinate of $Fx, Fy, Fz, M$ as Table 4 shows. If control command $F$ can accurately identify and map the foot movement directions, the operators can intuitively control the movements of the slave robot as they intended to. Thus, the prediction accuracy rate of the motion directions was regarded as a metric to identify intuitive mapping methods.

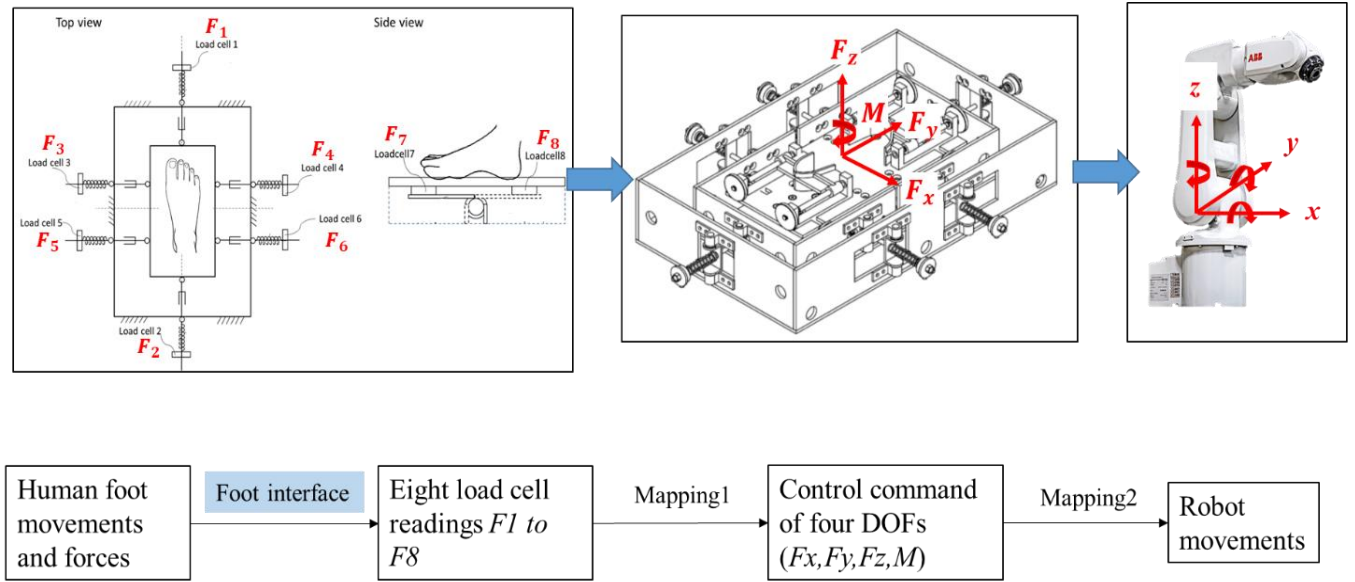

Figure 29: Mapping of foot interface system.

Table 4: Force coordinate system and motion directions

| Force | Fx- | Fx+ | Fy+ | Fy- | Fz+ | Fz- | M+ | M- |
|---|---|---|---|---|---|---|---|---|
| Direction | Left | Right | Forward | Backward | Toe up | Toe down | Supination | Pronation |
| Abbreviation | L | R | F | B | TU | TD | S | P |

The prediction accuracy can be calculated through:

$$Accuracy_{ij} = \frac{P_{cij}}{P_{tij}} \times 100\%$$

Where $j$ corresponds to subjects, $j = 1, 2 \ldots 10$, and $i$ corresponds to directions. For eight single axis directions, $i = 1, 2 \ldots 8$ correspond to {L, R, F, B, TU, TD, S, P}. For twelve diagonal test directions, $i = 1, 2 \ldots 12$ corresponding to the directions {LF, RF, LB, RB, LTU, RTU, LTD, RTD, FTU, BTU,

FTD, BTD}. The numerator $P_{cij}$ is the number of correct force vectors satisfying the requirements of the *i*th direction for *j*th subject in one motion trial. The conditions need to satisfy for a specific direction are listed in Table 5 and Table 6. The denominator $P_{tij}$ is the total number of all non-zero data vector in the *i*th direction for *j*th subject in one motion trial. For example, when calculate the accuracy rate $Acc_{11}$ of subject 1 in left direction in some mapping method. The $P_{c11}$ will be the number of force vector satisfy conditions "$F_x < 0 \ \& \ F_y = 0 \ \& \ F_z = 0 \ \& \ M = 0$" which regarded as the correct sub-movements in left direction. $P_{t11}$ is the number of all movements which represented by all the non-zero force vectors which satisfying "$F_x \neq 0$ or $F_y \neq 0$ or $F_z \neq 0$ or $M \neq 0$".

**Table 5: Prediction accuracy criterion for single axis directions**

| Directions | Abbreviation | Conditions |
|---|---|---|
| Left | L | *Fx<0; Fy = 0;Fz=0;M=0* |
| Right | R | *Fx>0; Fy = 0;Fz=0;M=0* |
| Forward | F | *Fx=0; Fy > 0;Fz=0;M=0* |
| Backward | B | *Fx=0; Fy < 0;Fz=0;M=0* |
| Toe up | TU | *Fx=0; Fy = 0;Fz>0;M=0* |
| Toe down | TD | *Fx=0; Fy = 0;Fz<0;M=0* |
| Supination | S | *Fx=0; Fy = 0;Fz=0;M>0* |
| Pronation | P | *Fx=0; Fy = 0;Fz=0;M<0* |

**Table 6: Prediction accuracy criterion for diagonal axis directions**

| Directions | Abbreviation | Conditions |
|---|---|---|
| Left &Forward | LF | *Fx < 0; Fy > 0;Fz=0;M=0* |
| Right & Forward | RF | *Fx > 0; Fy > 0;Fz=0;M=0* |
| Left & Backward | LB | *Fx < 0; Fy < 0;Fz=0;M=0* |
| Right & Backward | RB | *Fx > 0; Fy < 0;Fz = 0; M=0* |
| Left & Toe up | LTU | *Fx < 0; Fy = 0;Fz > 0;M=0* |
| Right & Toe up | RTU | *Fx > 0; Fy = 0;Fz > 0;M=0* |
| Left & Toe down | LTD | *Fx < 0; Fy = 0;Fz < 0;M=0* |
| Right & Toe down | RTD | *Fx > 0; Fy = 0;Fz < 0;M=0* |
| Forward & Toe up | FTU | *Fx = 0; Fy > 0;Fz > 0;M=0* |
| Backward & Toe up | BTU | *Fx = 0; Fy < 0;Fz > 0;M=0* |
| Forward & Toe down | FTD | *Fx = 0; Fy > 0;Fz < 0;M=0* |
| Backward & Toe down | BTD | *Fx = 0; Fy < 0;Fz < 0;M=0* |

The statics modeling and ICA method are tested using metric of direction prediction accuracy we defined above. One set of eight single axis directions' data were used to identify the mapping model when using the ICA. The diagonal data were also tested as the modeling data, the result was shown in Appendix A4 which are significantly lower than the single axis modeling's result. The statics modeling method uses result from kinematics calculations, thus there is no need to build mapping model in advance. Another sets of data in directions of single and diagonal axes were used to test these two methods Figure 30.

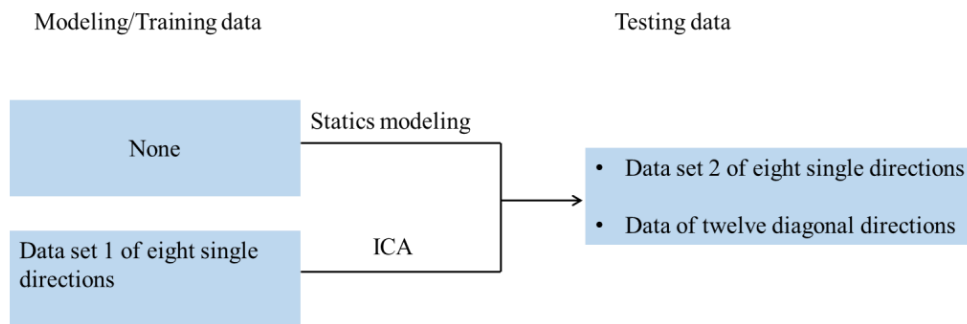

**Figure 30: Modeling/ training and testing data for two mapping methods.**

It is important to note that defining a zero region is necessary before checking the direction detection accuracy. As described in Section 2.5, our device has a well-defined home position. However, the pedal may not return exactly to this position due to the residual force exerted by the foot and motor noise, thus it is preferable to define a zero region instead of exact home position. Zero regions are chosen for each control signals ($F_x$, $F_y$, $F_z$ and $M$). For example, when the pedal moves forward, only the forward direction movements are the signal need to be identifed, other directions are regarded as noise, thus $F_x$, $F_z$ and $M$ are filtered to zero. The values of noise of $F_x$ $F_y$ $F_z$ and $M$ are selected using a minmax measure of subject specific data, as a too small range leads to a poor filtering result, and a too large range will reduce the measurement sensitivity. Figure 31 shows the *Fx* profile with time in single axis motions. The zero regions are represented in the form of a dash line band which covers most of the noise but does not affect the signals.

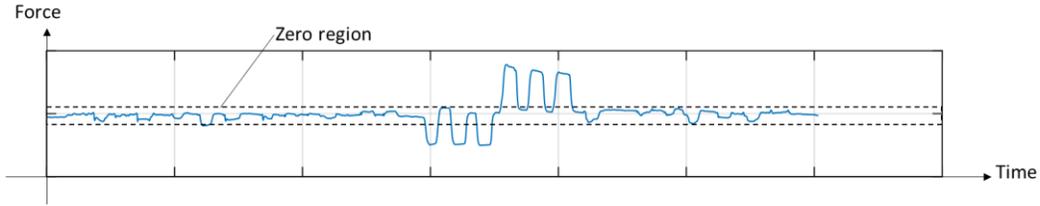

**Figure 31:** *Fx* **Force profile with time of in single axis motions.**

## 4.3 Result of statics modeling method

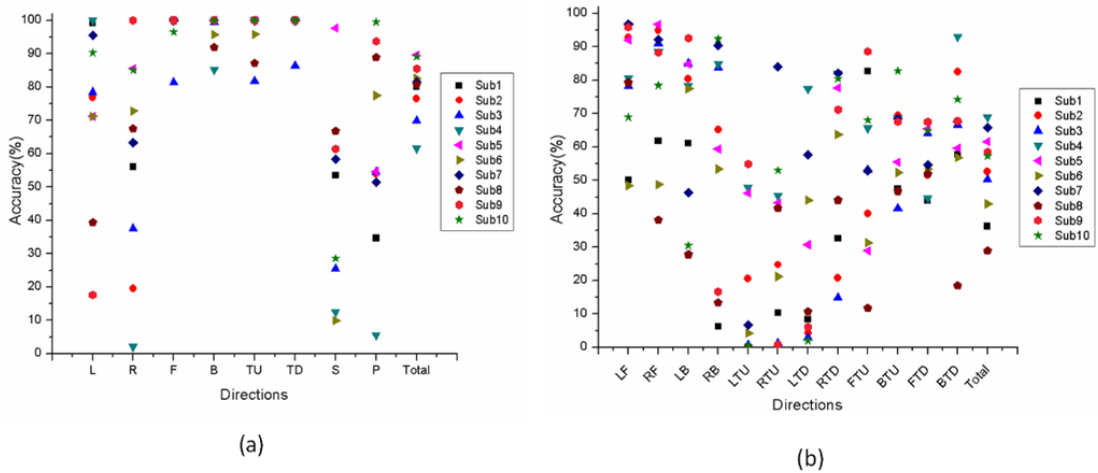

**Figure 32: Statics calculation method accuracy result in (a) single axis directions and (b) diagonal axis directions.**

Using the static modeling as mapping method yields 80% and 52% of correct selections results for single and diagonal directions respectively. In Figure 32(a), the detection accuracy is very high in the directions of F, B, TU and TD, which have the average detection accuracy of 96%, 96%, 97% and 99%. However, for the directions of L, R, S and P, the average detection accuracies fall to 70%, 62%, 50% and 65% respectively. Particularly, for some subjects the accuracy falls to less than 10%. The L/R translational motions and lateral-medial axial rotations of S/P interfere with each other and cannot be separated well. Those four directions will be quite difficult to control for the users if using the statics modeling method.

Figure 32 (b) shows the prediction result of diagonal motion test. The twelve diagonal directions comprise 4 sets of combinations of $Fx$ (±) and $Fy$ (±), $Fx$ (±) and $Fz$ (±), $Fy$ (±) and $Fz$ (±) (Figure 28). The average accuracy rate of these three groups are 70%, 33%, 58%. The accuracy rate of $Fx$-$Fy$ plane diagonal motion is relatively high, which is as expected since it is relatively easy for humans to control the foot movements in translational directions. However, it is a little harder to combine TU/TD rotations together with translation movements. The average total accuracy of twelve diagonal directions is just 52%, particularly, in directions of LTU and LTD, with accuracy of 18% and 24% respectively. The result in previous section already proves that four diagonal directions in $Fx$-$Fy$ plane have better accuracy, smaller direction derivation and more smooth movements.

## 4.4 Result of independent component analysis (ICA) method

Another method providing a user specific mapping from the load cells signals to the robot control inputs consists of using the independent component analysis (ICA). Based on different classifications of input data, we can separate the ICA method into global ICA and local ICA (Figure 33). In the global ICA method, eight load cells' readings $F1$ to $F8$ are input data and the expected output are four dof' motion commands $Fx, Fy, Fz$ and $M$ in the format of four independent components. In the local ICA method, the input signals of eight load cells are separated to two groups and built two sub ICA models respectively. The input signals of first group generated from load cells 3 to 6 are used to extract components $Fx$ and $M$; the inputs of loadcells 1, 2, 7 and 8 are used to separate components $Fy$ and $Fz$.

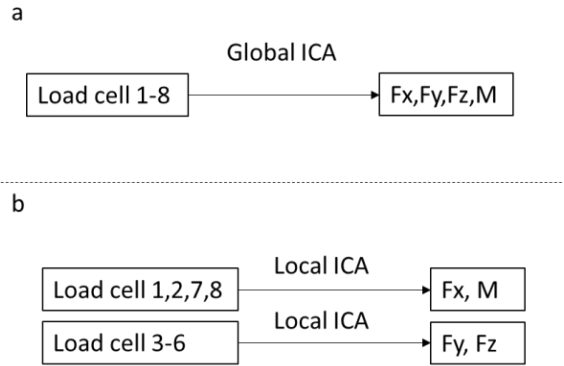

**Figure 33: Schematic diagram of (a) global ICA method (b) local ICA method.**

## 4.4.1 Global ICA

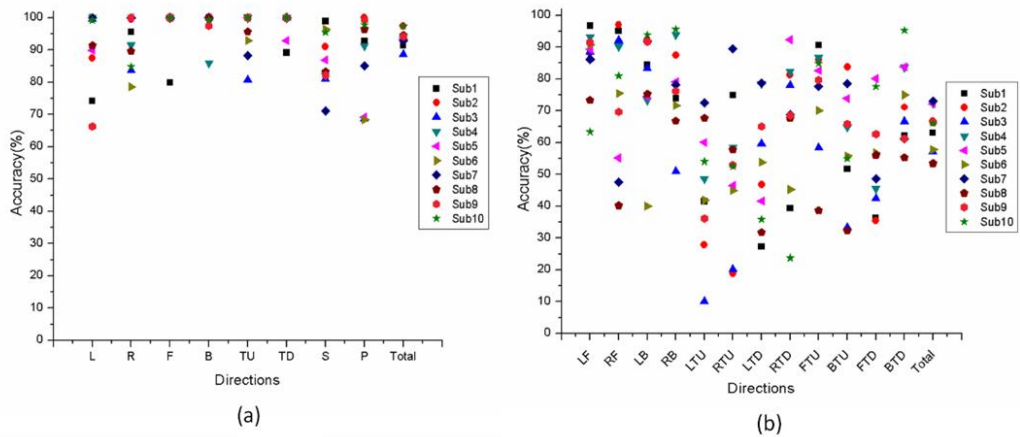

**Figure 34: Global ICA method. Accuracy prediction result on (a) Eight single directions (b) Twelve planar diagonal directions.**

Figure 34 (a) shows the average prediction accuracy over 10 subjects through using global ICA method in eight single directions. The result exhibits high accuracy in all directions with the average and standard deviation of 93%±4%. The accuracy of {F, B, TU, TD} maintain in high level with 98%, 97%, 96% and 98% respectively. The direction of {L, R, S, P} also have a good prediction accuracy with 91.64%, 91.45%, 88.03% and 88.05% which manifest significant improvements compared to the statics modeling method that are 69.73%, 61.87%, 50.33% and 64.97%.

The total average accuracy on twelve diagonal directions is 65% (Figure 34 (b)). The $Fx$-$Fz$ plane combinations still have relatively low accuracy same with the result in statics calculation method, especially in LTU direction which is 46%. However, the combined direction of LF has an accuracy rate of 86% which is highest among all diagonal directions. The $Fx$-$Fy$, $Fx$-$Fz$, $Fy$-$Fz$ planes have average accuracies of 79%, 54% and 65%. These values are higher compared to those using the statics modeling method which are 69%, 33%, 58%.

### 4.4.2 Local ICA

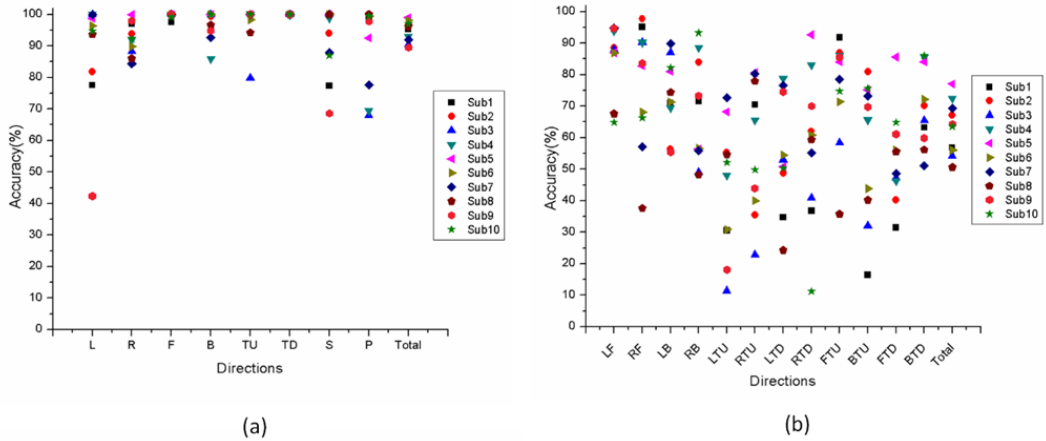

**Figure 35: Local ICA method (a) Eight single directions (b) Twelve diagonal directions**

The results when using the local ICA method are shown in Figure 35, exhibiting similar performance to those obtained with the global ICA method. The detection accuracies are all above 90% for eight directions of single axis motion. And the total average accuracy reaches 95% (Figure 35 (a)), with toe down direction achieving 100% accuracy! Even the lowest direction, left direction, has prediction accuracy of 90%. The total average accuracy on twelve diagonal directions is 63% (Figure 35 (b)) with lowest accuracy of 44% in LTU direction and highest accuracy 86% in LF direction.

Compare the results of single axes motions in statics modeling and ICA, the prediction accuracies in global and local ICA improved 13% and 14% relative to statics modeling in average total accuracy. For directions of F, B, TU and TD which statics modeling already has a good result, ICA maintain the high levels and have slightly better performance. In addition, it is worth mentioning that the two ICA methods can effectively separate directions of L/R and S/P which is difficult for static modeling. The average accuracy of L and R directions increased from 65% in statics modeling to 91% and 91.5% in global and local ICA. The average accuracy of S and P also has an obvious growth from 57% to 90.5% and 92.5%. When it comes to the result of diagonal directions. The ICA methods still exhibit better performance than statics modeling. The average total detection accuracy in diagonal direction increases by around 10% from 52%. The lowest accuracy in LTU direction increased from 18% in static modeling to 46% in global ICA and 44% in local ICA. However, the performance diversity of twelve diagonal directions are relatively big compared to single axis directions with standard deviation of 18%, 13% and 12%. Directions {LTU, RTU, LTD} show lower accuracy in both methods of statics modeling and ICA. To sum up, the two ICA methods show higher prediction accuracy than the statics modeling method in both single motion directions and diagonal directions. The methods of global ICA and local ICA exhibit similar prediction performance.

# Chapter 5 Conclusions and future work

## 5.1 Conclusion

Through reviewing the current human-machine interfaces in robotic surgery and focusing especially on the foot interfaces, it was found that existing foot interfaces are mainly composed of switches, pedals and buttons and limited to three dof. These interfaces present shortcomings such as a lack of haptic feedback, and offer only restricted discrete speeds and directions. With such interfaces, the operator may need to visually check the foot posture frequently, making the control more complex and causing fatigue. Therefore, we have designed a passive and compliant foot interface system which presents the following novelties:

(1) Our foot interface system can be used to control four dof of a robotic arm. The movement is not limited to a single discrete direction but in contrast can yield any combination of directions in four dof.

(2) The spring's network (with six compression springs and two torsion springs) forms a compliant system which can provide a feedback in force and position via the elastic interaction between the pedal and the operator's foot.

(3) The elastic-isometric model avoids the ceiling effect of input signals. When the motion reaches the geometric limits, the corresponding force sensor can still record the increasing force yielding a continuous transition between the elastic input and isometric input modes.

(4) The interface has an automatic positioning mechanism comprising multiple springs yielding a single minimum of elastic energy. Once the surgeon finishes operation movements and releases the pedal, the interface will go back to this neutral position automatically, thus providing a resting position for the operator's foot.

(5) The interface software has a built-in technique and algorithm to extract operator specific commands based on natural foot movement patterns thus enabling intuitive and efficient control of a slave robot.

## 5.2 Future work

A serial of experiments will be conducted to prove the concept and explore the human foot capability and the structure of the hardware and the algorithm of the system will be optimized according to the results of those user studies.

### 5.2.1 Comparison experiment and user study

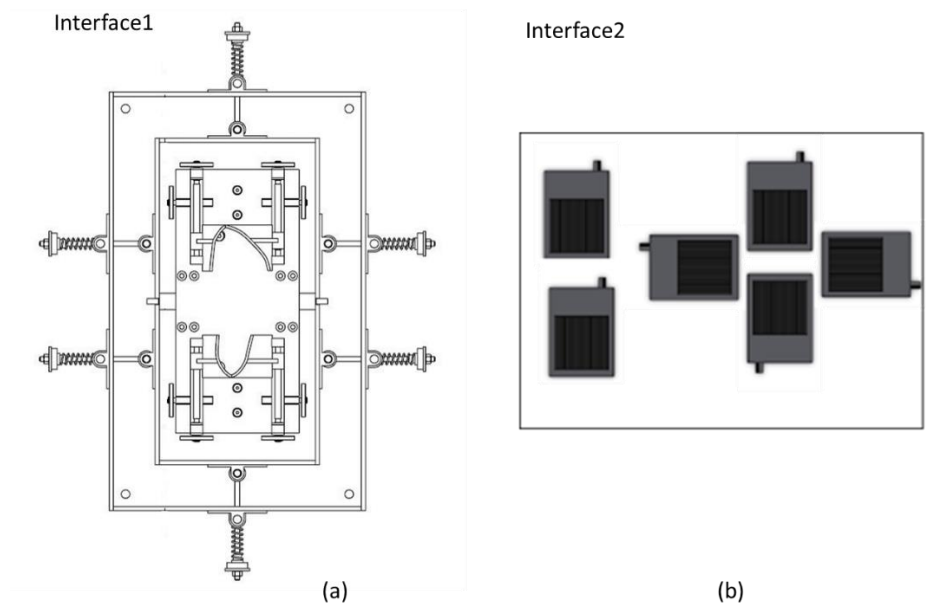

**Figure 36:Top view of (a) the compliant foot interface for pseudo-haptics (b) a button-type foot interface for 3 DOF control.**

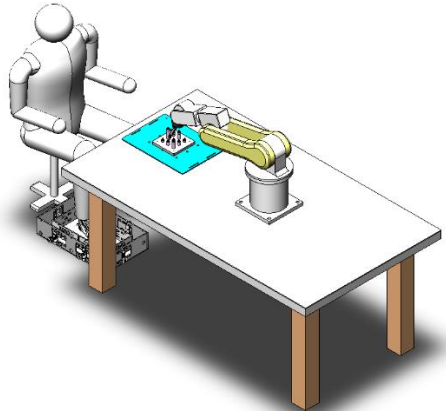

**Figure 37:Comparison experiment set up.**

The current existing foot interface are in button type which have several buttons on the planar platform. The advantage of using a foot pedal interface in compare to buttons has not been proven yet. Thus, to prove our interface are superior than the conventional interface in intuitiveness, efficiency and accuracy. A comparison experiment will be designed and conducted.

First, the foot interface teleoperation system will be built including a slave robot (ABB irb120, ABB, Pte. Ltd). A comparison button type foot interface will be designed as Figure 36(b) shows. The both foot interfaces will act as master interfaces and be compared in experiments with naïve operators for the teleoperation of the robotic arm in skillful manipulation. A control system with motor modeling and parameter tuning and simulation will be built to achieve smooth velocity or force control in the teleoperation system.

During the experiment, the principles of *i)* the different effects with designed compliant interface relative to a rigid foot interface; *ii)* multiple axes combinations; *iii)* the necessary pseudo-haptic feedback for surgery will be developed and tested. The detailed experiment protocol is listed in Appendix A7. The experiment platform for this experiment has already built up. The experiment protocol has already approved by IRB of NTU in 3$^{rd}$ of July 2018.

## 5.2.2 Surgical scenario experiments and user study

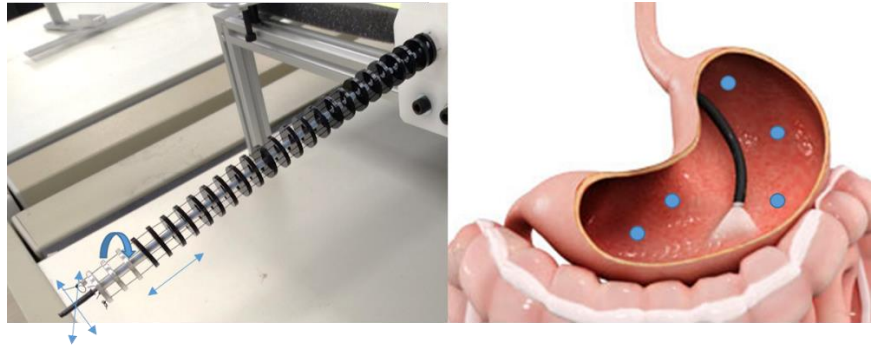

**Figure 38: Surgical application experiment. (a) Flexible endoscope (b) Stomach model with marked target points**.

In a second phase, the foot interface system will be integrated with a dedicated surgical robotic assistive system. Its potential for clinical application will be evaluated in the application of robotic endoscopic diagnose. The testing experiment will be conducted with a flexible robot with four dof which will be manufactured following the principles of tendon-driven mechanism[75][76]. The rough prototype of the flexible robot is shown in Figure 38(a). Then mechanism will be actuated by tendons which are connect to motors though pulleys. Each dof needs two tendons with two motors. The experiment will simulate a scenario of medical endoscope diagnose. The subject will tele-control the movements of the flexible robot with endoscope camera at the tip though foot interface. The camera provided a view of the unknown environment (e.g. human stomach cavity) with marked targets (e.g tumor in human body) inside. The subject is required to teleoperate the flexible robot to find and reach the hidden targets (Figure 38(b)). The task completion time, endoscope movement information, foot motion information will be recorded. This experiment will not only limit to study foot but also the foot-eye coordination. As the endoscope camera view is different with the human natural view, a visual motor test and study of human [82] in endoscopic camera view will be needed.

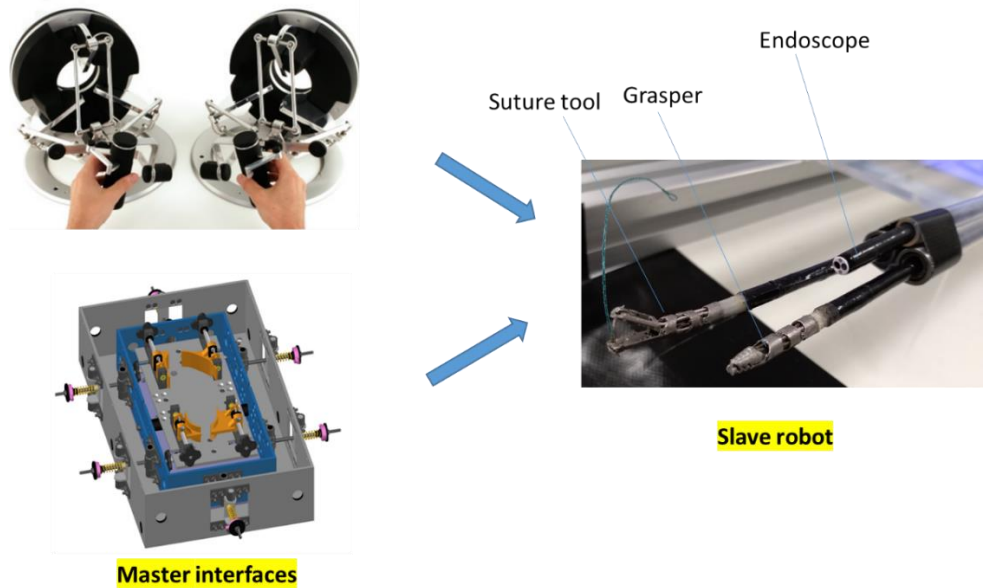

**Figure 39: Master and slave interfaces in robotic surgical system.**

Thirdly, the cooperation between the foot acting as a third arm and two natural hands will be tested in some surgical tasks. It is still an open question that what kinds of tasks are suitable for hand-foot coordination and what's the capability of the cooperation can achieve. There are a lot of surgical cases which need three tools operate together. For example, in the cooperative suturing task [80], the surgeon needs to use grasper to hold the tissue and suture tool to do the suturing, at the same time, the endoscope needs to move from front view to back view to check whether and when the suture needle puncture the tissue. Except the endoscope, the foot interface will also be tested to control the other surgical tools. In some surgical operations such as appendectomies, cholecystectomies, and splenectomies, the camera do not require to be frequently repositioned [5]. The camera can be fixed, the movement and grasping of the forceps which used to holding the tissue can be controlled by human foot through the foot interface and cooperate one hand/two hands in surgical tasks. The haptic feedback of grasping will be provided using pseudo haptics by displaying the effect of grasping force on the monitor.

A more comprehensive robotic surgical platform with hand consoles and foot interface will be built in the future. In which, two haptic devices of Omega7 from Force Dimension Inc will be used for hands' teleoperation on two surgical tools. Each Omega7 can control 6 dof movements plus one dof of grasping which is dexterous enough for hands operation using position control. The hands gesture will map to position of surgical tool in scaling. The developed foot interface will be used to control assisted tool of the endoscope /forceps through rate control. Coordination and control capability among three surgical tools controlled by two hands and one foot will be tested in the specific surgical task like suturing.

### 5.2.3 Optimization of structure and methodology

The experiments we introduced in the previous sections serve for exploring the key aspects e.g. foot force, motion, mapping etc. that can be used to determine an intuitive, robust foot interface to control surgical robot efficiently and accurately. Then after the experiments and analysi, accordingly, the foot interface will be optimized in the following aspects:

(1) The optimization of force field of foot interface to best facilitate human foot operation will be conducted based on the result of experiment. Currently, springs are used as passive actuators in the foot interface to provide haptic feedbacks. The parameters like the location of springs, the stiffness of springs, the motion range of springs etc. are all critical issues to build an optimized force field which effectively changed with foot motion. The active actuator may be used to some joints to make the force more controllable.

(2) The parallel-serial structure has been used for our interface. This structure may not be the most efficient way to reflect human foot motions. The structure optimization will be studied.

(3) The subject-specific mapping method with ICA was currently selected to reflect the users' intention. It has shown its superiority in single Cartesian direction prediction. To build a complete methodology of foot control, optimization of the mapping between natural foot motion to control commands still need to study deeper through analyze of characteristic of human motion.

(4) Furthermore, identify best mapping algorithm from force control command to robot motions. The linear relationship and force-rate mapping was currently employed. There are many alternatives as position-rate, force-force, etc. and non-linear relations. These alternatives should be qualifiedly and quantitively analyzed and summarized with an optimal control combination according to simulation and experiment result.

(5) The foot interface system should allow two-way communication with feedback path. Although many studies were conducted on haptics and there are mature hand haptic devices, how to effectively and efficiently provide multi-dof haptic feedback to foot is still unknown. The questions like which part of foot is more suitable for receiving haptic feedback. What kinds of haptic feedback is more effective for foot (e.g. vibration, force, tactile); does pseudo-haptics work well with visual assistance in robotic surgery etc. will be explored in the future study.

# Reference

[1] M. Nurok, T. M. Sundt, and A. Frankel, "Teamwork and Communication in the Operating Room: Relationship to Discrete Outcomes and Research Challenges," *Anesthesiol. Clin.*, vol. 29, no. 1, pp. 1–11, 2011.

[2] L. Lingard, R. Reznick, S. Espin, G. Regehr, and I. DeVito, "Team Communications in the Operating Room," *Acad. Med.*, vol. 77, no. 3, pp. 232–237, 2002.

[3] S. Elprama, K. Kilpi, P. Duysburgh, A. Jacobs, L. Vermeulen, and J. Van Looy, "Identifying barriers in telesurgery by studying current team practices in robot-assisted surgery," *Proc. ICTs Improv. Patients Rehabil. Res. Tech.*, 2013.

[4] "https://www.materprivate.ie/dublin/centre-services/all-services/robotic-surgery/." .

[5] S. Kim and S. C. Lee, "Technical and instrumental prerequisites for single-port laparoscopic solo surgery : state of art," vol. 21, no. 15, pp. 4440–4446, 2015.

[6] H. Mohrmann-Lendla and A. G. Fleischer, "The effect of a moving background on aimed hand movements," *Ergonomics*, vol. 34, no. 3, pp. 353–364, 1991.

[7] S. S. Kommu, P. Rimington, C. Anderson, and A. Rané, "Initial experience with the EndoAssist camera-holding robot in laparoscopic urological surgery," *J. Robot. Surg.*, vol. 1, no. 2, pp. 133–137, 2007.

[8] J. Stolzenburg, T. Franz, M. Do, K. Turner, and E. Liatsikos, "COMPARISON OF THE FREEHAND (R) ROBOTIC CAMERA HOLDER TO HUMAN ASSISTANTS DURING THE ENDOSCOPIC EXTRAPERITONEAL RADICAL PROSTATECTOMY (EERPE)-A PROSPECTIVE RANDOMIZED STUDY OF 50 CASES," *J. Endourol.*, vol. 23, pp. A3–A3, 2009.

[9] S. N. Yulun Wang, Goleta, Darrin Uecker, "Speech Interface For An Automated Endoscopic System US 6,463,361 B1," 2002.

[10] R. Polet and J. Donnez, "Using a laparoscope manipulator (LAPMAN) in laparoscopic gynecological surgery.," *Surg. Technol. Int.*, vol. 17, pp. 187–91, 2008.

[11]  G. F. Buess *et al.*, "A new remote-controlled endoscope positioning system for endoscopic solo surgery: The FIPS endoarm," *Surg. Endosc.*, vol. 14, no. 4, pp. 395–399, 2000.

[12]  D. P. Noonan, G. P. Mylonas, J. Shang, C. J. Payne, A. Darzi, and G. Z. Yang, "Gaze contingent control for an articulated mechatronic laparoscope," *2010 3rd IEEE RAS EMBS Int. Conf. Biomed. Robot. Biomechatronics, BioRob 2010*, pp. 759–764, 2010.

[13]  C. Staub, S. Can, B. Jensen, A. Knoll, and S. Kohlbecher, "Human-computer interfaces for interaction with surgical tools in robotic surgery," *Proc. IEEE RAS EMBS Int. Conf. Biomed. Robot. Biomechatronics*, pp. 81–86, 2012.

[14]  "News From the Food and Drug Administration," vol. 318, no. 13, p. 2017, 2017.

[15]  L. N. S. Andreasen Struijk, "An inductive tongue computer interface for control of computers and assistive devices," *IEEE Trans. Biomed. Eng.*, vol. 53, no. 12, pp. 2594–2597, 2006.

[16]  D. Johansen, C. Cipriani, D. B. Popovic, and L. N. S. A. Struijk, "Control of a Robotic Hand Using a Tongue Control System-A Prosthesis Application," *IEEE Trans. Biomed. Eng.*, vol. 63, no. 7, pp. 1368–1376, 2016.

[17]  I. Farkhatdinov, N. Roehri, and E. Burdet, "Anticipatory detection of turning in humans for intuitive control of robotic mobility assistance," *Bioinspiration and Biomimetics*, vol. 12, no. 5, 2017.

[18]  P. Doctoral and E. N. Robotique, "Supernumerary Robotic Arm for Three-Handed Surgical Application: Behavioral Study and Design of Human-Machine Interface Elahe ABDI," vol. 7343, 2017.

[19]  E. Abdi, E. Burdet, M. Bouri, S. Himidan, and H. Bleuler, "In a demanding task, three-handed manipulation is preferred to two-handed manipulation," *Sci. Rep.*, vol. 6, pp. 1–11, 2016.

[20]  E. Abdi, E. Burdet, M. Bouri, and H. Bleuler, "Control of a supernumerary robotic hand by foot: An experimental study in virtual reality," *PLoS One*, vol. 10, no. 7, pp. 1–15, 2015.

[21]  F. R. Jalan *et al.*, "Basic Anatomy of the Foot," *Head Neck*, vol. 77, no. 4, pp. 1–21, 2013.

[22]  E. (Lancaster U. Velloso, D. (Hasso P. I. Schmidt, J. (Lancaster U. Alexander, H. (Lancaster U. Gellersen, and A. (Max P. I. for I. Bulling, "The Feet in Human – Computer Interaction : A Survey of Foot-Based Interaction," *ACM Comput. Surv.*, vol. 48, no. 2, pp. 1–35, 2015.

[23]  K. Zhong, F. Tian, and H. Wang, "Foot menu: Using heel rotation information for menu selection," *Proc. - Int. Symp. Wearable Comput. ISWC*, no. State 1, pp. 115–116, 2011.

[24]  L. Zhou, W. Meng, C. Z. Lu, Q. Liu, Q. Ai, and S. Q. Xie, "Bio-Inspired Design and Iterative Feedback Tuning Control of a Wearable Ankle Rehabilitation Robot," *J. Comput. Inf. Sci. Eng.*, vol. 16, no. 4, p. 041003, 2016.

[25]  D. Carrera, "3D Modeling Of Human Knee And Movement Simulation," 2014.

[26]  V. Paelke, C. Reimann, and D. Stichling, "Foot-based mobile Interaction with Games," *Ace'04*, pp. 321–324, 2004.

[27]  J. Abascal, S. Barbosa, M. Fetter, T. Gross, P. Palanque, and M. Winckler, *Human-computer interaction – INTERACT 2015: 15th IFIP TC 13 international conference Bamberg, Germany, september 14-18, 2015 proceedings, Part III*, vol. 9298. 2015.

[28]  S. W. Alderson, "Electrically Operated Artificial Arm For Above-The-Elbow Amputees," 1952.

[29]  M. C. Carrozza *et al.*, "A wearable biomechatronic interface for controlling robots with voluntary foot movements," *IEEE/ASME Trans. Mechatronics*, vol. 12, no. 1, pp. 1–11, 2007.

[30]  A. L. Simeone, E. Velloso, J. Alexander, and H. Gellersen, "Feet movement in desktop 3D interaction," *IEEE Symp. 3D User Interfaces 2014, 3DUI 2014 - Proc.*, pp. 71–74, 2014.

[31]  J. Alexander, T. Han, W. Judd, P. Irani, and S. Subramanian, "Putting your best foot forward," *Proc. 2012 ACM Annu. Conf. Hum. Factors Comput. Syst. - CHI '12*, p. 1229, 2012.

[32]  K. M. Stanney, *Handbook of Virtual Environments*. 2002.

[33] T. Pakkanen and R. Raisamo, "Appropriateness of foot interaction for non-accurate spatial tasks," *Conf. Hum. Factors Comput. Syst.*, pp. 1123–1226, 2004.

[34] T. Han, J. Alexander, A. Karnik, P. Irani, and S. Subramanian, "Kick: Investigating the use of Kick Gestures for Mobile Interactions," *Proc. 13th Int. Conf. Hum. Comput. Interact. with Mob. Devices Serv.*, pp. 29–32, 2011.

[35] A. J. Sellen, G. P. Kurtenbach, and W. A. S. Buxton, "The Prevention of Mode Errors Through Sensory Feedback," *Human-Computer Interact.*, vol. 7, no. 2, pp. 141–164, 1992.

[36] D. W. Podbielski, J. Noble, H. S. Gill, M. Sit, and W. C. Lam, "A comparison of hand- and foot-activated surgical tools in simulated ophthalmic surgery," *Can. J. Ophthalmol.*, vol. 47, no. 5, pp. 414–417, 2012.

[37] M. A. van Veelen, J. J. Jakimowicz, and G. Kazemier, "Improved physical ergonomics of palaroscopic surgery," *Minim. Invasive Ther. Allied Technol.*, vol. 13, no. 3, pp. 161–166, 2004.

[38] S. H. Kim and D. B. Kaber, "Design and evaluation of dynamic text-editing methods using foot pedals," *Int. J. Ind. Ergon.*, vol. 39, no. 2, pp. 358–365, 2009.

[39] M. Azmandian, M. Hancock, H. Benko, E. Ofek, and A. D. Wilson, "Haptic Retargeting: Dynamic Repurposing of Passive Haptics for Enhanced Virtual Reality Experiences," *Proc. 2016 CHI Conf. Hum. Factors Comput. Syst. - CHI '16*, pp. 1968–1979, 2016.

[40] P. Tommasino, A. Melendez-Calderon, E. Burdet, and D. Campolo, "Motor adaptation with passive machines: A first study on the effect of real and virtual stiffness," *Comput. Methods Programs Biomed.*, vol. 116, no. 2, pp. 145–155, 2014.

[41] M. Mace, P. Rinne, J.-L. Liardon, P. Bentley, and E. Burdet, "Comparison of isokinetic and isometric handgrip control during a feed-forward visual tracking task," *Proc. Int. Conf. Rehabil. Robot.*, pp. 792–797, 2015.

[42] L. S. G. L. Wauben, M. A. Van Veelen, D. Gossot, and R. H. M. Goossens, "Application of ergonomic guidelines during minimally invasive surgery: A questionnaire survey of 284 surgeons," *Surg. Endosc. Other Interv. Tech.*, vol. 20, no. 8, pp. 1268–1274, 2006.

[43]  M. A. Van Veelen, C. J. Snijders, E. Van Leeuwen, R. H. M. Goossens, and G. Kazemier, "Improvement of foot pedals used during surgery based on new ergonomic guidelines," *Surg. Endosc. Other Interv. Tech.*, vol. 17, no. 7, pp. 1086–1091, 2003.

[44]  B. Hatscher, M. Luz, and C. Hansen, "Foot Interaction Concepts to Support Radiological Interventions," vol. 17, no. September, pp. 3–13, 2017.

[45]  S. Mazzoleni, J. Van Vaerenbergh, E. Stokes, G. Fazekas, P. Dario, and E. Guglielmelli, "An ergonomic modular foot platform for isometric force/torque measurements in poststroke functional assessment: A pilot study," *J. Rehabil. Res. Dev.*, vol. 49, no. 6, p. 949, 2012.

[46]  J. M. Gilbert, "The EndoAssist[TM] robotic camera holder as an aid to the introduction of laparoscopic colorectal surgery," *Ann. R. Coll. Surg. Engl.*, vol. 91, no. 5, pp. 389–393, 2009.

[47]  B. Paul Moraviec, "Surgical Mechanism Control System US 9,639,953 B2," US 9,639,953 B2, 2017.

[48]  M. O. Schurr *et al.*, "Trocar and instrument positioning system TISKA: An assist device for endoscopic solo surgery," *Surg. Endosc.*, vol. 13, no. 5, pp. 528–531, 1999.

[49]  R. G. Thorlakson, "Method And Foot Pedal Apparatus For Operating A Microscope US 5,787,760," 1998.

[50]  M. Michael E. Metzler, Eden Prairie, MN(US); Merlin Hall, Wildwood, "Foot Controller Including Multiple Switch Arrangement With Heel Operated, Door-type Switch Actuator US 6,689,975 B2," 2004.

[51]  A. F.Mora, "Foot Actuated Switch US 8076599 B2."

[52]  J. L. Elkins, "Foot-operated Controller US 7,186,270 B2," 2007.

[53]  H. H. Randal P. Goldberg, Michael Hanuschik, "Ergonomic Surgeon Control Console In Robotic Surgical Systems US8,120,301 B2," 2012.

[54]  M. G. Munro, "Automated laparoscope positioner: Preliminary experience," *Am. Assoc. Gynecol. Laparosc.*, vol. 1, no. 1, pp. 67–70, 1993.

[55] A. Minor, R. Ordorica, J. Villalobos, and M. Galan, "Device to provide intuitive assistance in laparoscope holding," *Ann. Biomed. Eng.*, vol. 37, no. 3, pp. 643–649, 2009.

[56] J. A. Peter Berkelman, Philippe Cinquin, Alain Jacquet, "System For Positioning On A Patient An Observation And/Or Intervention Device US 8,591,397 B2," 2013.

[57] B. B. Clement VIDAL, "System For Controlling Displacement Of An Interention Device US 2017/0303957 A1."

[58] S. Voros, G. P. Haber, J. F. Menudet, J. A. Long, and P. Cinquin, "ViKY robotic scope holder: Initial clinical experience and preliminary results using instrument tracking," *IEEE/ASME Trans. Mechatronics*, vol. 15, no. 6, pp. 879–886, 2010.

[59] "http://www.hiwin.tw/products/me/mtg_h100.aspxpa." .

[60] A. Mirbagheri, F. Farahmand, A. Meghdari, and F. Karimian, "Design and development of an effective low-cost robotic cameraman for laparoscopic surgery: RoboLens," *Sci. Iran.*, vol. 18, no. 1 B, pp. 59–71, 2011.

[61] A. Mirbagheri *et al.*, "Operation and human clinical trials of robolens: an assistant robot for laparoscopic surgery," *Front. Biomed. Technol.*, vol. 2, no. 3, pp. 184–190, 2015.

[62] Y. Wang, R. U. S. A. Data, P. E. B. Davis, and S. Taylor, "Automated Endoscope System For Optimal Positioning US 5,515,478," 1996.

[63] T. Kawai, M. Fukunishi, A. Nishikawa, Y. Nishizawa, and T. Nakamura, "Hands-free interface for surgical procedures based on foot movement patterns," *2014 36th Annu. Int. Conf. IEEE Eng. Med. Biol. Soc. EMBC 2014*, pp. 345–348, 2014.

[64] J. Y. K. Chan *et al.*, "Foot-controlled robotic-enabled endoscope holder for endoscopic sinus surgery: A cadaveric feasibility study," *Laryngoscope*, vol. 126, no. 3, pp. 566–569, 2016.

[65] X. Dai, B. Zhao, Y. He, Y. Sun, and Y. Hu, "A foot-controlled interface for endoscope holder in functional endoscopic sinus surgery," *Front. Biomed. Devices, BIOMED - 2017 Des. Med. Devices Conf. DMD 2017*, pp. 2017–2018, 2017.

[66] E. Abdi and J. Olivier, "Foot-Controlled Endoscope Positioner for Laparoscopy : Development of the Master and Slave Interfaces," pp. 111–115, 2016.

[67] I. Díaz, J. J. Gil, and M. Louredo, "A haptic pedal for surgery assistance," *Comput. Methods Programs Biomed.*, vol. 116, no. 2, pp. 97–104, 2014.

[68] A. Pusch and A. Lécuyer, "Pseudo-haptics: from the theoretical foundations to practical system design guidelines," *Proc. 13th Int. Conf. multimodal interfaces*, pp. 57–64, 2011.

[69] A. Lécuyer, "Simulating Haptic Feedback Using Vision: A Survey of Research and Applications of Pseudo-Haptic Feedback," *Presence Teleoperators Virtual Environ.*, vol. 18, no. 1, pp. 39–53, 2009.

[70] M. Achibet, A. Girard, A. Talvas, M. Marchal, and A. Lecuyer, "Elastic-Arm: Human-scale passive haptic feedback for augmenting interaction and perception in virtual environments," *2015 IEEE Virtual Real. Conf. VR 2015 - Proc.*, pp. 63–68, 2015.

[71] L. Dominjon, J. Perret, and A. Lécuyer, "Novel devices and interaction techniques for human-scale haptics," *Vis. Comput.*, vol. 23, no. 4, pp. 257–266, 2007.

[72] A. Lécuyer, J.-M. Burkhardt, and L. Etienne, "Feeling bumps and holes without a haptic interface: the perception of pseudo-haptic textures," *Proc. SIGCHI Conf. Hum. Factors Comput. Syst. - CHI '04*, vol. 6, no. JANUARY 2004, pp. 239–246, 2004.

[73] I. Kim, K. Tadano, T. Kanno, and K. Kawashima, "Implementing pseudo haptic feedback in a semi-isometric master interface for robotic surgery," *Int. J. Adv. Robot. Syst.*, vol. 14, no. 5, pp. 1–9, 2017.

[74] R. L. Barnett, "Foot controls: Riding the pedal," *Ergon. Open J.*, vol. 2, pp. 13–16, 2009.

[75] M. Kouchi, "Foot Dimensions and Foot Generation Shape: and Differences Ethnic Origin Due," *Anthropol. Sci.*, vol. 106, no. Supplement, pp. 161–188, 1998.

[76] J.-P. Merlet, *Parallel Robots*. 2001.

[77] J. Rastegar, "M A N I P U L A T O R WORKSPACE ANALYSIS USING THE N " V ~ , Y2 Yi X 1 Xl," vol. 25, no. 2, pp. 233–239, 1990.

[78] F. Gao, W. Li, X. Zhao, Z. Jin, and H. Zhao, "New kinematic structures for 2-, 3-, 4-, and 5-DOF parallel manipulator designs," *Mech. Mach. Theory*, vol. 37, no. 11, pp. 1395–1411, 2002.

[79] D. Surdilovic, J. Radojicic, and N. Bremer, *Cable-Driven Parallel Robots*, vol. 32. 2015.

[80] J. M. Tartaglia *et al.*, "TENDON-DRIVEN ENDOSCOPE AND METHODS OF INSERTION US6858005 B2," 2005.

[81] Y. H. Lee and J. J. Lee, "Modeling of the dynamics of tendon-driven robotic mechanisms with flexible tendons," *Mech. Mach. Theory*, vol. 38, no. 12, pp. 1431–1447, 2003.

[82] D. Formica, D. Campolo, F. Taffoni, F. Keller, and E. Guglielmelli, "Motor adaptation during redundant tasks with the wrist," *Proc. Annu. Int. Conf. IEEE Eng. Med. Biol. Soc. EMBS*, pp. 4046–4049, 2011.

[83] J. M. Sackier and Y. Wang, "Surgical Endoscopy From Concept to Development," *Surg. Endosc.*, vol. 8, no. 8, pp. 63–66, 1994.

[84] J. Velotta, J. Weyer, A. Ramirez, J. Winstead, and F. V. R. F. Vl, "Vilas-Boas, Machado, Kim, Veloso (eds.)," vol. 11, pp. 1035–1038, 2011.

[85] N. Jennings, J. Nakamura, S. Reed, and A. Wilson, "Determining Leg Dominance Using the Unipedal Stance Test ( UPST )," pp. 9–10, 2015.

# Appendix

## A0 Comparison table between existing foot interface and the developed one

Table A1: Comparison between current available technologies of foot interfaces in robotic surgery and the developed foot interface

| Product Name | Endex [54] | PMASS [55] | AESOP [62][83] | ViKy [56][57][58] | RoboLens [60][61] | HIWIN [59] | Da Vinci [53] | The developed foot interface |
|---|---|---|---|---|---|---|---|---|
| DOFs | 1 | 1 | 3 | 3 | 3 | 3 | - | 4 |
| Control mode | Single axis | Single axis | Two combined axes; Pressure proportional speed | Single axis; Constant speed | Single axis; Constant speed | Single axis; Constant speed | Switches | Fully combined in 4 DOF |
| Positioning and resting | × | × | × | × | × | ✓ | × | ✓ |
| Force feedback | × | × | × | × | × | × | × | ✓ |
| Force measurement | rigid | rigid | rigid | rigid | rigid | rigid | rigid | compliant |
| Foot adaptation | × | × | × | × | × | × | ✓ | ✓ |
| Subject specific mapping | × | × | × | × | × | × | × | ✓ |

× means the system don't have such mechanism; ✓ means the system have related mechanism/technique.

# A1 Foot interface assembly

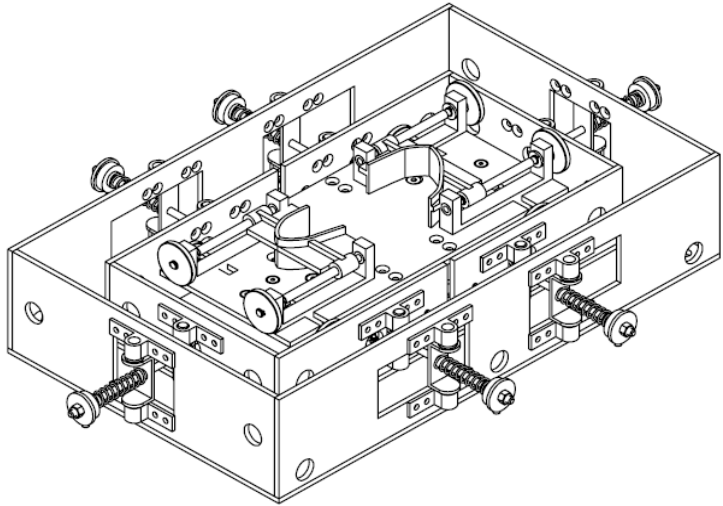

**Figure A1: Perspective view of the developed foot interface**

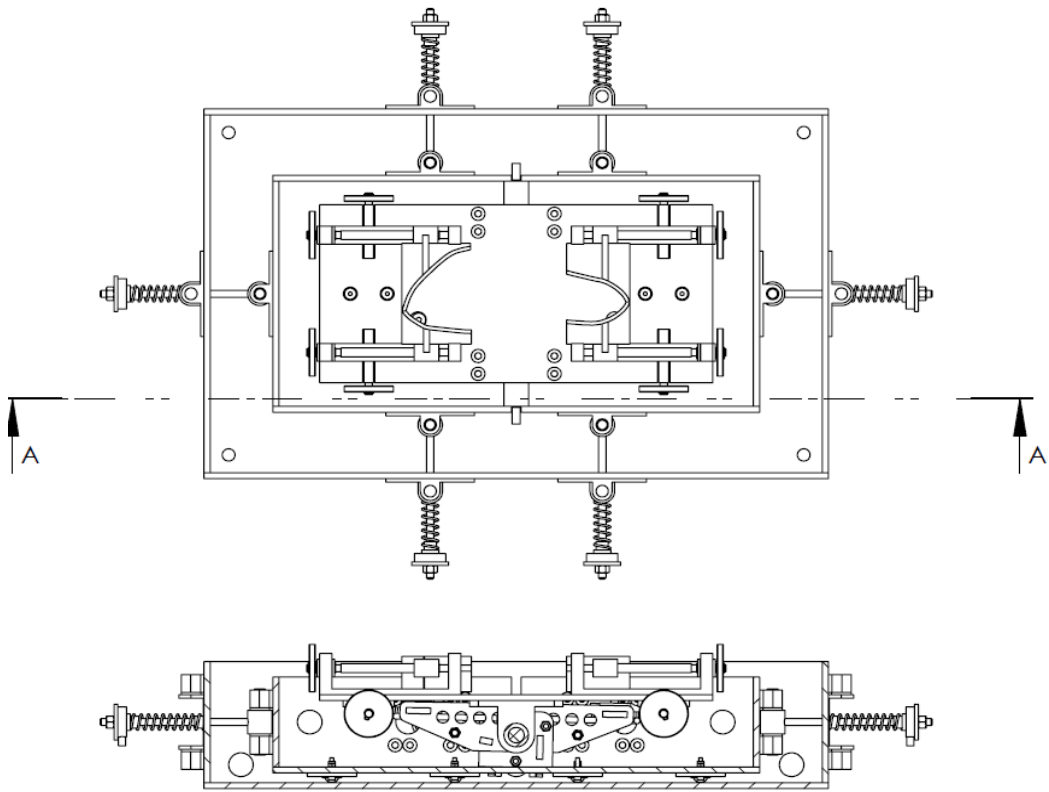

**Figure A2: Top and cross-section views of the foot interface.**

Perspective view

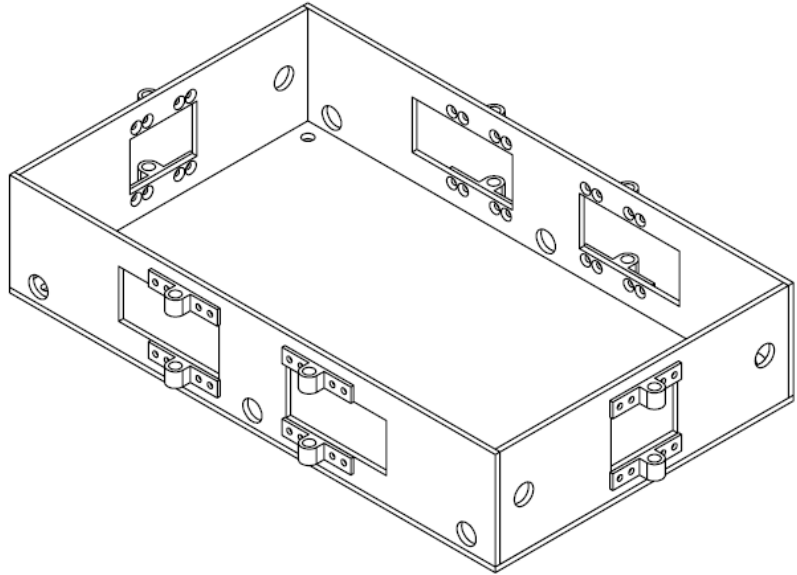

**Figure A 3: Perspective view of the base assembly of the foot interface.**

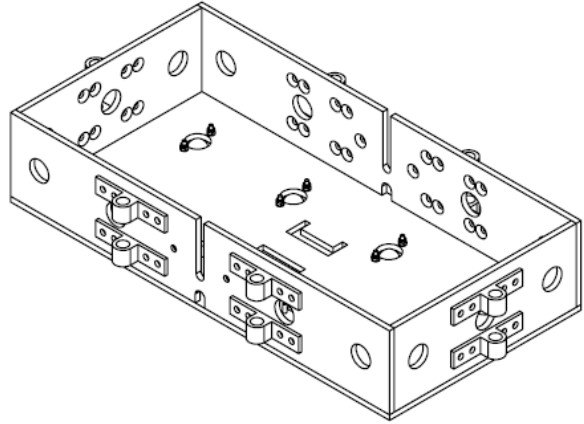

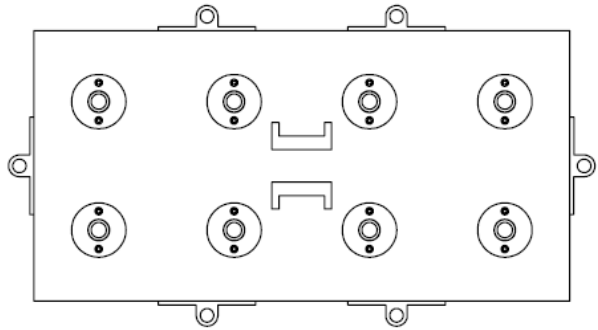

**Figure A4: perspective and bottom views of the mobile frame of pedal assembly.**

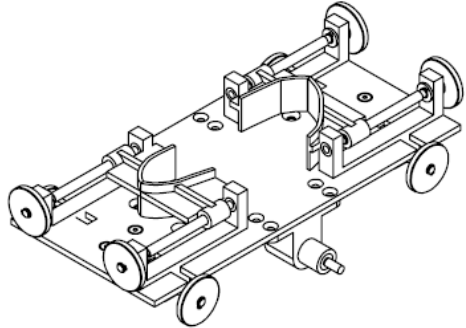

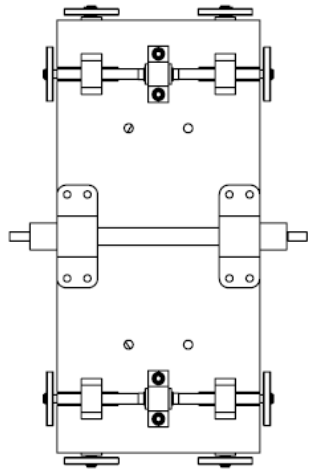

**Figure A5: Perspective and bottom views of the treadle plate with adjustable foot fastening mechanism of the pedal assembly.**

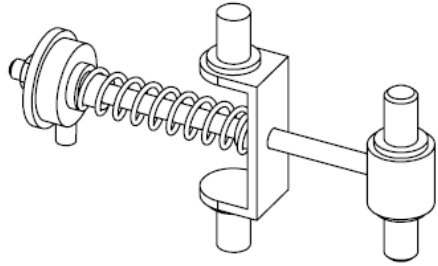

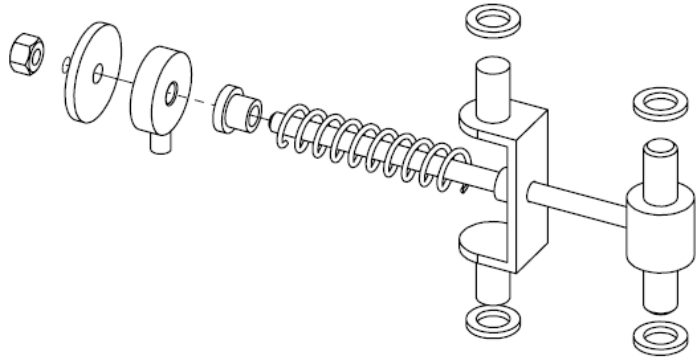

**Figure A6: Perspective and explosive views of the compression spring assembly of the foot interface.**

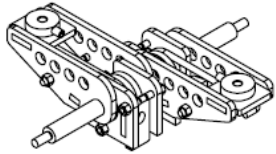

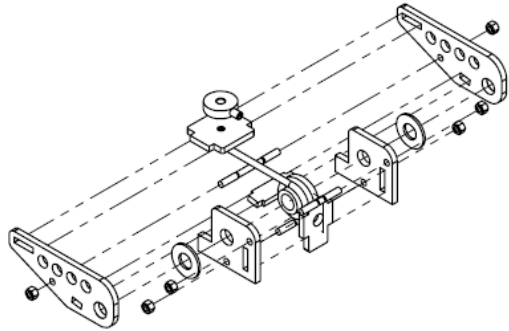

**Figure A7: Perspective and explosive views of the torsion spring assembly of the foot interface.**

## A2 k-nearest neighbor's algorithm (kNN) method

K-Nearest Neighbors (kNN), a simple machine learning algorithm, can be used to identify operator specific foot moving directions. With this method, each direction is assigned a class label, e.g. 1 for the forward, 2 for the backward, etc. Then motion data with related labels about directions will be input to kNN model as training data. After the training is completed, new data can be tested to identify the direction labels using this kNN model.

Figure A8 shows the prediction result of eight single directions when using a kNN classification method. The first figure is the detection accuracy when using load cells reading *F1* to *F8* as training data (case 1); the second figure is the result when using kinematics calculation result *Fx, Fy, Fz* and *M* as training data (case 2).

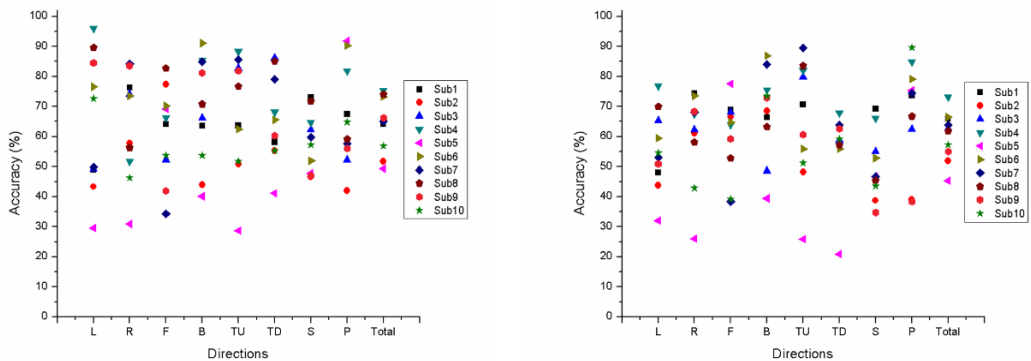

**Figure A8: kNN method (a) using F1 to F8 as training and test data (b) using *Fx,Fy,Fz* and M as training and testing data.**

The total average prediction rate for case 1 is 64%, this is slightly better but similar as that in case 2 which has accuracy of 60%. The average accuracy for each direction for these two cases are shown in the following Table A. The supination direction has the lowest accuracy with 58% and 49% in these two cases. The backward direction has the highest accuracy rate for case 1 with 68%. Case 2 have the same number but in pronation direction.

Table A4: Accuracy rate via kNN for different training data.

| Directions /Accuracy (%) | Training data | |
| --- | --- | --- |
| | $F1$ to $F8$ (case 1) | $Fx, Fy, Fz$ and $M$ (case 2) |
| L | 64 | 55 |
| R | 63 | 60 |
| F | 61 | 59 |
| B | 68 | 67 |
| TU | 67 | 64 |
| TD | 65 | 56 |
| S | 58 | 49 |
| P | 66 | 68 |
| Total | 64 | 60 |
| Standard deviation | 3 | 6 |

As a whole, the performance of using $F1$ to $F8$ as training data is slightly better than using $Fx, Fy, Fz$ and $M$ as training data. However, comparing the result of kNN case 1 and statics mapping, kNN method exhibits relative low performance in prediction results. The total average accuracy of statics method was 80% and this number was only 64% in kNN case 1. The detection accuracy in the directions of forward/backward and toe up/toe down reached 96% and 98% in statics modeling, by contrast, these are just 63% and 60% in kNN case 1. The statics method didn't perform well in directions of left/right and supination/pronation with prediction accuracy of 66% and 58%. And kNN also presented low performance with 58% and 59% in those directions. Given the above comparison, the static modeling method exhibits much better prediction results than the kNN method.

# A3 Experiment protocol of comparison between developed foot interface and existing commercialized foot interface

We have developed a passive four-degree-of-freedom foot interface which can be used by an operator to control a robotic arm was developed. The experiment is made to compare this foot interface (Foot interface 1) with the common commercialized foot interface, button-type foot interface (Foot interface 2), on movement capability, motion efficiency, positioning accuracy, and intuitiveness to user.

## A3.1 Experiment set up

The Figure A9 shows the top views of the two kinds of foot interfaces. The left one is the interface we developed and the right one is the button-type foot interface with six buttons/switches. The number of buttons will increase with the degree of freedom we are going to control. With four DOFs movements, eight buttons will be needed.  As the existing foot interface is limited to at most three DOFs, we have made a six-button foot interface as a reference foot interface as Figure A9 (b) shows (Foot interface 2).

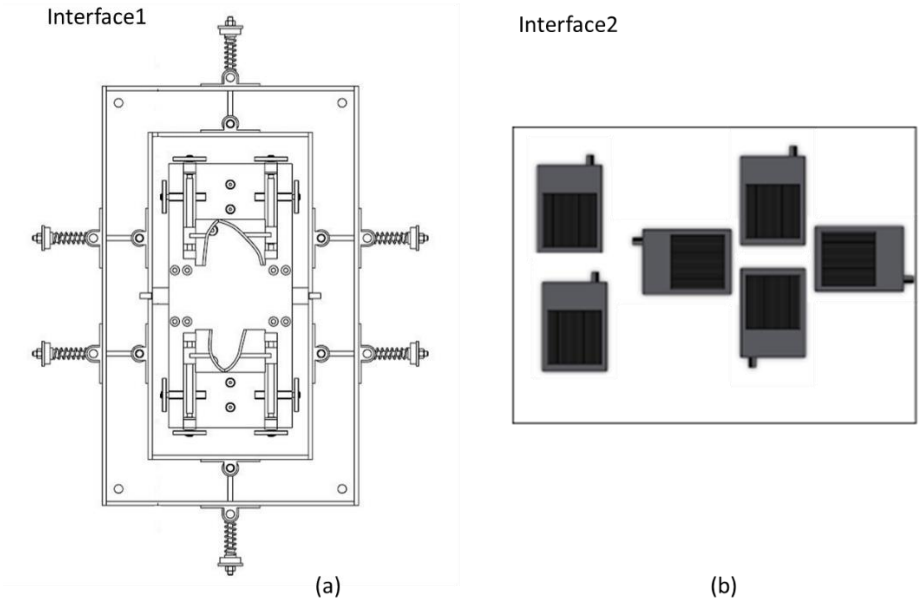

**Figure A9: Top view of (a) the compliant foot interface for pseudo-haptics (b) a button-type foot interface for 3 DOF control.**

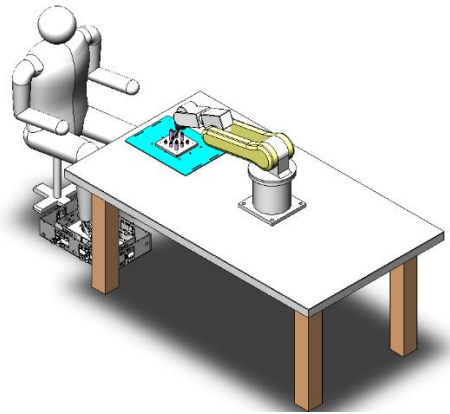

**Figure A 10: Experiment set up.**

The above foot interfaces can act as master device to tele-operate a slave robot. Figure A 10 shows the experiment set up with foot interface1. The ABB IRB120 robotic arm is used in our experiment as a slave robot. Some related parameters of the robot are listed in the Table .

**Table A5: Some parameters of ABB IRB 120 robot.**

| Feature/Performance | |
|---|---|
| Acceleration time 0-1m/s | 0.07s |
| Position repeatability | 0.01mm |
| Reach | 580mm |

Ten subjects (four female) will be involved in the experiment. Each subject will conduct two tasks with both two interfaces. The sequence of using the two interfaces are random. Half of the subjects will start with interface 1.

Each participant's trials consist the following phases:

(1) Dominant leg identification phase: Dominant leg will be determined through a ball kicking test. (first day)

(2) Calibration and habituation phase: calibration data will be collected through using interface 1, similar practice will also be conducted through interface 2.  (first day)

(3) Task phase:

   1) Task1 with ABB robotic arm controlled by interfaces 1(morning), interface 2 (afternoon). (second day)

   2) Task2 with ABB robotic arm controlled by interface 1 (morning), interface 2 (afternoon). (third day)

(4) Test phase: same tasks with task phase. (fourth day)

**A3.2 Dominant leg identification phase**

Before involved in the calibration and task phase, we will do a test to identify the dominant leg for each subject. The dominant leg tends to have bigger muscles and used for fine motor skills, e.g. striking the soccer ball. The non-dominant leg is the one used as the balance leg. There are many common tests to determine the dominant leg/foot. For example, ask the subject to kick a ball, take a step forward [84] or keep balance with one leg stand [85]. We are going to use the method of ball kicking test, as kicking a ball is quite similar with operating a foot pedal.

**Starting position**

The starting position for the participant will be marked with a piece of tape on the floor. The subjects' feet will be placed at hip width apart and parallel to each other. The ball will be placed on marked position in the middle front of the subject to the forward direction.

**Procedure**

First, the subject is requested to kick the ball to a target area which is three meters away. Three trials will be conducted for each subject. Their preferred legs will be observed and recorded which are regarded as their dominant legs. Next, the subject is asked to answer that which leg he/she prefer to use as dominant leg.

The dominant leg will be determined through the ball kick test, their answer will act as a reference. if the subject's answer is different with the result of the test, another backup test (A.3) will be used to that subject. This test has the same weight with the previous ball kick test.

**A3.3 Calibration and Habituation Phase**

The calibration and habituation phase are also arranged at a first day for subject familiar with the two interfaces and collecting calibration data. In this phase, both interfaces are tested in multiple single directions movements' control.

At first, the subject is requested to move the pedal of interface 1 to eight directions, namely forward, backward, left, right, toe up, toe down, supination and pronation. The details of each direction are described in Section 4.1.1. In each movement, the pedal started at the neutral position and then moved to the extreme position. Three movements in each direction will be conducted. In each movement, extreme position will be kept for 1 second.

All the movements are conducted without viewing. The motion force data will be collected and used to build specific calibration model for each subject.

Secondly, the subject is requested to conduct the six directions' motions via interface 2. As the foot interface 2 are designed with six buttons which represent six discrete directions {forward, backward, left, right, up, down}, the subject presses those buttons sequentially. Then repeat three times for each group of movements.

After this phase, the subject will get familiar with the two interfaces and its related responding DOF and controlled directions.

### A3.4 Task phase

The task phase will be conducted <u>on the second and third day.</u> It consists of the execution of the two tasks, described in A8.4.1 and A8.4.2. Before each task, the subject can put his foot comfortably in the interface and try a few movements <u>for a few minutes.</u> When using the foot interface 1, the subject is requested to step the dominant foot on the pedal. The pedal will stay in the neutral position of the foot interface in the original state. For the button interface, the subject will be requested to use only the dominant leg. The non-dominant leg will be separated from the buttons through a partition board in order to prevent it for activating the buttons. In addition, the two tasks will be asked to complete in a blind state. If the subject has to check with visual assistance, <u>the times and duration of eye check will be recorded through pressing a button by the experimental observer</u>.

### A3.4.1 Space positioning task

This task tests the space motion control capability of the foot interface through tele-controlling the robotic arm to reach different space-points.

**Material:**

Task board with 5 pegs of different heights (60mm, 40mm, 35mm, 25mm, and 20mm) is used in this experiment. Each peg equipped with a LED light button (diameter: 12mm) on the top of it. The dimensions of the task board and pegs are determined in accordance with the commercialized laparoscopic training box.

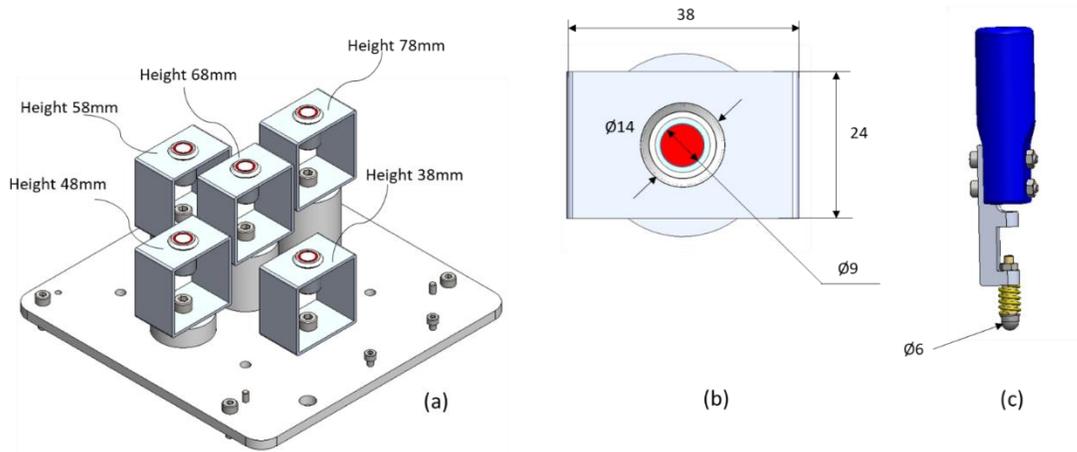

**Figure A11: Task1 board dimensions. (a) Perspective view of task board. (b) Top view of one peg with LED button. (c) Robot end effector pressing tool.**

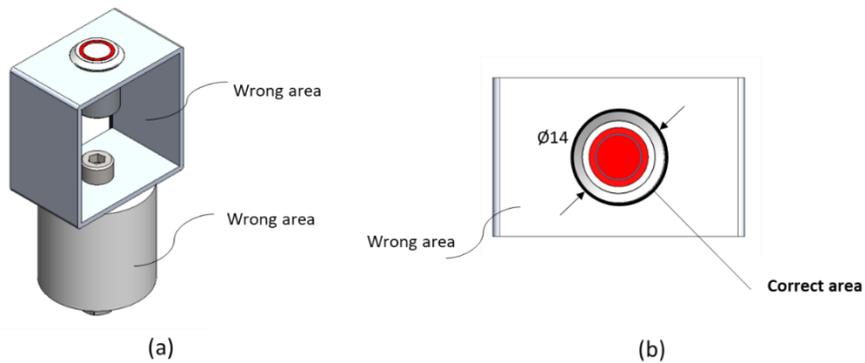

**Figure A 12: Wrong and correct area of task 1.**

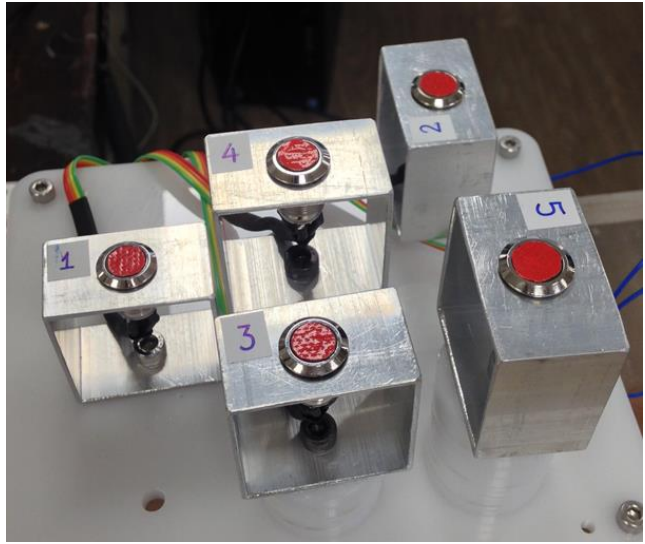

**Figure A13: Task1 board physical map.**

**Starting position**

The task board is placed on the platform near the robotic arm. There are 5 pegs with LED buttons, randomly located in the nine holes of the task board. There are two random arrangements. At the beginning, all the LED light are lightened/closed.

**Procedure**

The subjects were told to navigate the pressing tool to the position of a LED button, and presses the button, then moves to the next one, etc. The LED lights will be closed/lightened one by one as quickly as possible without touching wrong area shown Figure A 12. The task is completed when 5 LED buttons are all closed/lightened. It will be regard as successful trial when all the lights are lighted/closed without wrongly hitting the peg within 5 minutes. Once the trial fails, the subject will continue until the task is completed. The task has to be successfully executed **ten times.** The peg locations will be rearranged after five successful trial, and the same random sequence of configurations is used for all subjects. There will be a five minutes break after five successful trial. Successive successful trials are not required and there is no requirement on the order of the movements.

For each trial, the following data were collected by experiments:

(1) **Total completion time**: duration to complete the task. <u>The time is set as 0</u> when the first light is lightened/closed. <u>The time is set as end</u> when the last light is closed/lightened. The time can be recorded through Arduino clock.

(2) **Number/times of wrong robot movement**: the number of robot movements which hit the pegs. The wrong touching area and correct area are shown in **Error! Reference source not found.**. This number will be record by a button manually activated by the experiment conductor.

(3) **Number of eye check and its duration**: This number will be record by a button manually activated by the experiment conductor.

(4) **The robot movement trajectory.**

(4) **Foot movement**: The foot movement and positions will be recorded by a motion tracking sensor attached on the subject's foot.

### A3.4.2 Path following task

This task is to move a ring along a coiled wire with minimal collisions between the ring and the wire. It could help testing precision and fine control of movement via foot interfaces.

**Material:**

Aluminum wire path: 120mm × 90mm/120mm*70mm (diameter: 2.5mm)

Tool: with ring of diameter 14 mm

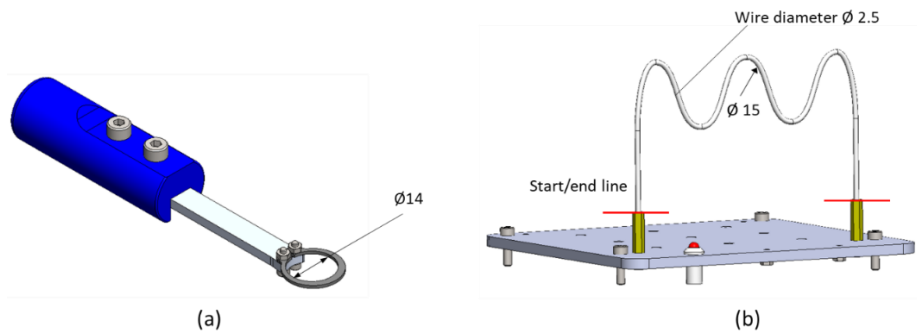

**Figure A14: Task2 materials. (a) Tool with ring at the end (b) Task board.**

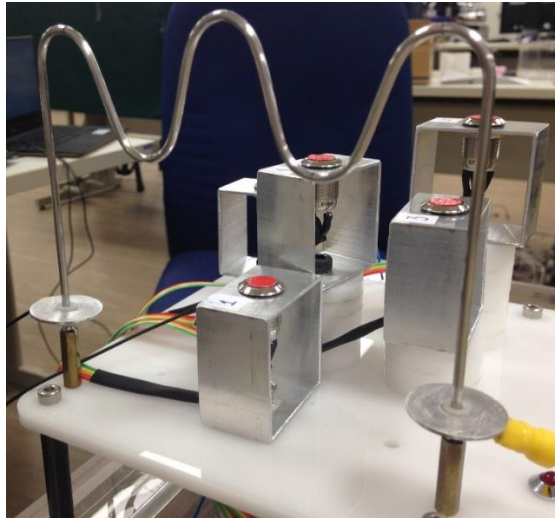

**Figure A15: Task2 board physical map.**

**Starting position**

The task board is placed on the platform near the robotic arm with planar curve wire mounted on it in a vertical position. The robotic arm connects with a tool that is assembled together with the ring. The ring is placed at the one side of the wire.

**Procedure**

The subject is requested to control the movements of the ring to follow the path of the wire with minimum number/time of collision between the wire and the ring. Once the ring contacts the wire, a red warning light will be lighted with buzz sound. The task is finished when the ring is moved from one side to the other side of the wire. Both sides are equipped with a conducive plate, once the ring touches the plate, the electrical circuit will be connected.   The task is required to be successfully executed **<u>five times</u>**. There will be a 5 minutes break after three successful trial. The task will continue even there is collision occur during the process. All trials should be completed. For each trial, the following data are collected during the experiment:

(1) **<u>Total completion time</u>**: duration of the whole task. The time starts to record when the ring moves out of the starting plate. And once the ring touches the finishing plate, the time stop to record. Both sides of plate can be start/end points.

(2) **<u>Number/time of collisions</u>**: The number of collisions between the ring and the wire during movements. The collisions will activate a buzzer and a LED light. The collision times can be recorded through buzzer or/and LED signals from Arduino board.

(3) **<u>Number of eye check and its duration:</u>** This number will be record by a button manually activated by the experiment conductor.

(4) **<u>The robot movement trajectory</u>**

(5) **<u>Foot movement</u>**

### A3.5 Test phase

A test phase which to check the learning behavior will be conducted on the fourth day. The given tasks are the same with the second and third day. The detailed operation procedures are described in Section A 8.4. The tasks are required to be successfully executed **<u>five times</u>** for each subject.

### A3.6 Experimental subjective questionnaire

At the end, the participates were given a questionnaire to record their experiences with respect to comfort, ease of use and difficulties in handling the system, as well as with the precision achieved with the device. The questionnaire is designed as follows:

Questionnaire (5 points scale: 1=strongly unappreciated; 2=unappreciated;3= undecided; 4 = appreciated; 5 = strongly appreciated)

Q1. The mapping between foot movements and the robot movements is intuitive.

strongly disagree, disagree, neutral, agree, strongly agree

Q2. It was easy for me to control the robot arm with the foot.

strongly disagree, disagree, neutral, agree, strongly agree

Q3. I could control the robot arm accurately with the foot

strongly disagree, disagree, neutral, agree, strongly agree

Q4: Did you meet some difficulties in the operation?

no difficulty, some difficulties, a lot of difficulties

If you had difficulties, please specify them:

Q5: It was physically tiring for me to operate the robot

not tiring, tiring, very tiring

Q6: It was mentally tiring for me to operate the robot

not tiring, tiring, very tiring